\documentclass{article}
\pdfoutput=1

\usepackage[preprint,nonatbib]{neurips_2023}

\usepackage[utf8]{inputenc} 
\usepackage[T1]{fontenc}    
\usepackage{hyperref}       
\usepackage{url}            
\usepackage{booktabs}       
\usepackage{amsfonts}       
\usepackage{nicefrac}       
\usepackage{microtype}      
\usepackage{xcolor}         

\usepackage{caption}
\usepackage{subcaption}
\usepackage{graphicx}
\usepackage{amsmath}

\usepackage{amsthm}
\usepackage{multirow}
\usepackage{booktabs}

\usepackage{algorithm}
\usepackage{algorithmicx}
\usepackage[noend]{algpseudocode}
\usepackage{amsmath, amsfonts, amssymb, amsthm, amsbsy, amscd, bm, bbm,mathrsfs}    
\usepackage{graphicx}
\usepackage{subcaption}
\usepackage{cleveref}
\usepackage{enumitem}
\setlist{leftmargin=5mm}
\usepackage{titlesec}
\usepackage{wrapfig}
\usepackage{multirow}

\usepackage{soul}
\newcommand{\Note}[1]{}
\renewcommand{\Note}[1]{\hl{[#1]}}

\graphicspath{{figs/}}

\numberwithin{equation}{section}

\newtheorem{othertheorem}{othertheorem}[section]
\newtheorem{lemma}[othertheorem]{Lemma}

\newtheorem{claim}[othertheorem]{Claim}

\theoremstyle{definition}
\newtheorem{definition}[othertheorem]{Definition}
\newtheorem{remark}[othertheorem]{Remark}

\theoremstyle{definition}

\renewcommand{\v}{\bm{v}}
\newcommand{\g}{\bm{g}}

\newcommand{\E}{\mathbb{E}}

\newcommand{\var}{\operatorname{Var}}

\newcommand{\eff}{{\rm eff}}
\newcommand{\dip}{{\rm dp}}

\colorlet{darkgreen}{green!65!black}
\colorlet{darkblue}{blue!75!black}
\colorlet{darkred}{red!80!black}
\definecolor{lightblue}{HTML}{0071bc}
\definecolor{lightgreen}{HTML}{39b54a}

\definecolor{shilv}{HTML}{16A951}

\title{DPFormer: Learning Differentially Private Transformer on Long-Tailed Data}

\author{%
  Youlong Ding\thanks{Part of work done while at WeBank AI.} \\
  Shenzhen University\\
  Shenzhen, China\\
  \texttt{dingyoulon@gmail.com} \\
  \And
  Xueyang Wu \\
  Hong Kong University\\of Science and Technology \\
  Hong Kong SAR, China\\
  \texttt{xwuba@connect.ust.hk} \\
  \AND
  Hao Wang \\
  Rutgers University \\
  Piscataway, NJ, USA\\
  \texttt{hw488@cs.rutgers.edu} \\
  \And
  Weike Pan \\
  Shenzhen University \\
  Shenzhen, China\\
  \texttt{panweike@szu.edu.cn} \\
}

\begin{document}

\maketitle

\begin{abstract}
The Transformer has emerged as a versatile and effective architecture with broad applications. However, it still remains an open problem how to efficiently train a Transformer model of high utility with differential privacy guarantees. In this paper, we identify two key challenges in learning differentially private Transformers, i.e., heavy computation overhead due to per-sample gradient clipping and unintentional attention distraction within the attention mechanism. In response, we propose DPFormer, equipped with Phantom Clipping and Re-Attention Mechanism, to address these challenges. Our theoretical analysis shows that DPFormer can reduce computational costs during gradient clipping and effectively mitigate attention distraction (which could obstruct the training process and lead to a significant performance drop, especially in the presence of long-tailed data). Such analysis is further corroborated by empirical results on two real-world datasets, demonstrating the efficiency and effectiveness of the proposed DPFormer.
\end{abstract}


\section{Introduction}
Differentially private deep learning has made remarkable strides, particularly in domains such as image classification~\cite{tramerdifferentially,golatkar2022mixed,de2022unlocking} and natural language processing~\cite{yu2022differentially,li2022large,he2023exploring}. This success can be largely attributed to the availability of extensive pre-trained models, offering robust and diverse foundations for further learning. However, such reliance on vast, pre-existing datasets poses a significant challenge when these resources are not accessible or relevant. This hurdle becomes particularly pronounced when it is necessary to train differentially private Transformers using only domain-specific data gathered from real-world scenarios rather than generic, large-scale datasets.

This issue is glaringly evident in tasks like sequence prediction, which are integral to a wide range of applications, such as commercial recommender systems. These systems are designed to model and predict user behavior based on sequences of interactions like clicks or purchases. The reliance on domain-specific sequence data coupled with the constraints of differential privacy can significantly compromise the performance of such systems, particularly when large-scale public datasets or pre-existing pre-trained models are not at our disposal.

The challenges posed by this scenario can be summarized into two key hurdles. The first one stems from the inherent nature of real-world data, which typically follows a long-tailed distribution, where a small fraction of data occur frequently, while a majority of data appear infrequently. This poses the intrinsic hardness of high-utility differentially private training, which, based on its sample complexity~\cite{dwork2009complexity}, necessitates a \emph{sufficiently} large volume of data to discern general patterns without resorting to the memorization of individual data points~\cite{carlini2019secret,feldman2020does}. Our theoretical analysis shows that during differentially private training of Transformers, attention scores tend to be skewed by long-tailed tokens (i.e., tokens with fewer occurrences), therefore leading to huge performance drops.

The second hurdle arises from the resource-intensiveness of deep learning with differential privacy, which is primarily due to the requirement of clipping per-sample gradient. This requirement not only complicates the learning process but also places a significant computational burden, especially when resources are limited or when scalability is a priority.

To address these issues, we propose DPFormer (Figure~\ref{fig:DPFormer}), a methodology for learning differentially private Transformers. One of our primary contributions is the introduction of an efficient technique, Phantom Clipping. This technique addresses the computational burden associated with differential privacy by computing the clipped gradient without the need for the per-sample gradient. This allows for the efficient scaling of the learning process.
Additionally, to further enhance model utility under differentially private training, we introduce the Re-Attention Mechanism, which aims to mitigate the attention distraction phenomenon, ensuring the model's focus is directed toward the most pertinent elements of the input. This enables effective learning, especially in the presence of long-tailed data.

Experiments on public real-world datasets demonstrate the efficiency and effectiveness of DPFormer. These results not only validate our approach but also demonstrate the practical applicability of our model in scenarios where data is limited and domain-specific, and differential privacy is a crucial requirement. 

\section{Related Work}
\textbf{Differentially Private Deep Learning}. 
The most related works are \cite{yu2022differentially,li2022large} which finetune Transformer-based language models with differential privacy, and  \cite{anil-etal-2022-large} which trains differentially private BERT~\cite{hochreiter1997long}, a masked language model built upon Transformer. Their success is largely attributed to large-scale public information, which may not be available in some scenarios.

\textbf{Effective Training of Transformer.} The training of the Transformer models can be notoriously
difficult, relying heavily on  layer normalization~\cite{ba2016layer}, model initialization~\cite{huang2020improving}, learning rate warmup~\cite{popel2018training}, etc.
Beyond applying existing techniques proposed in the context of non-private learning, one intriguing research question is: Can we develop effective Transformer-training techniques that are specialized for private training? In this paper, we make the first attempt and propose the Re-Attention Mechanism, as discussed in Section~\ref{sec:reattn}.

\textbf{Differenital Private Learning on Heavy-Tailed Data}. 
Previous work has also considered the differentially private learning algorithms on heavy-tailed data~\cite{wang2020differentially,hu2022high,pmlr-v162-kamath22a}. This line of research is mainly concerned with differential private stochastic optimization (DP-SCO). Note that the notion of heavy-tailed there is different from the focus of this work. As pointed out in~\cite{pmlr-v162-kamath22a}, the setting they actually consider is dealing with heavy-tailed gradients due to unbounded values in the input data. 

For detailed analysis about the related work, please refer to appendix~\ref{apdx:related}.

\section{Preliminaries}

\textbf{Problem Setting: Sequential Prediction\footnote{This setting does not compromise universality. Transformer trained for the sequential prediction can be viewed as the \emph{universal} model for sequence modeling, enabling other tasks (e.g., classification) through fine-tuning.}.} 
Since Transformers are designed to predict the next token in an autoregressive manner, in this paper, we focus our evaluation on sequential prediction tasks, where each training sample consists of a sequence of tokens\footnote{In this paper, we will use `token' to denote the discrete unit within the input sequence and `vocabulary size' to represent the total count of relevant entities, generalizing their definitions associated with language modeling.} Given the preceding tokens $[s_1, s_2, ..., s_{t-1}]$, the task is to predict the next token ${s}_t$. 
Note that in practice (as is also the case for all our datasets), training data is typically long-tailed, in the sense that a small number of tokens occur quite frequently while others have fewer occurrences. Our goal is to train a Transformer with DP-SGD~\cite{AbadiCGMMT016}
such that it can predict the next token accurately while preserving differential privacy. 

\begin{definition}
\textbf{$(\varepsilon, \delta)$-Differential Privacy (DP)~\cite{dwork2006calibrating,dwork2014algorithmic}:} A randomized mechanism $\mathcal{M}:\mathcal{D}\rightarrow \mathcal{R}$  satisfies $(\varepsilon, \delta)$-differential privacy if for any two datasets $\mathcal{D}, \mathcal{D}'\in {\rm Domain}(\mathcal{M})$ that differ in one record and for all $S \in {\rm Range(\mathcal{M})}$ it holds that $\operatorname{Pr}(\mathcal{M}(\mathcal{D}) \in \mathcal{S}) \leq e^{\varepsilon} \operatorname{Pr}\left(\mathcal{M}(\mathcal{D}') \in \mathcal{S}\right)+\delta$.
\end{definition}
One desirable property of DP is that it ensures privacy (in terms of $\varepsilon$ and $\delta$) under composition.
Based on this property, DP-SGD~\cite{AbadiCGMMT016} injects calibrated Gaussian noise into model gradients in each training step to achieve differential privacy as follows,
\begin{equation}
\begin{split}
    \bm{G} &= \frac{1}{B}\left(\sum_{i=1}^B \g_i\cdot \operatorname{Clip_C}(\|\g_i\| + \sigma_{\dip} \cdot \mathcal{N}(0, \mathbf{I}))\right),
\end{split}
\label{eq:DPSGD}
\end{equation}
where $\g_i$ is the gradient of the $i$-th sample in the minibatch of size $B$. $C$ is the clipping norm, $\operatorname{Clip_C}(\|g_i\|)=\operatorname{min}(C/\|g_i\|, 1)$, ensuring that the sensitivity of the averaged gradient $G$ is bounded by $\Delta_G \leq \|g_i\cdot \operatorname{Clip}(\|g_i\|) \| \leq C$. $\dip$ is the noise multiplier derived from privacy accounting tools~\cite{balle2018privacy,wang2019subsampled}.

In the following sections, we introduce two novel and significant techniques of DPFormer, \textbf{Phantom Clipping} and \textbf{Re-Attention Mechanism}, which provide more efficient and precise DP-enhanced Transformer modeling.

\begin{figure}[h]
\centering
\begin{subfigure}{0.7\textwidth}
\centering
\includegraphics[width=9cm]{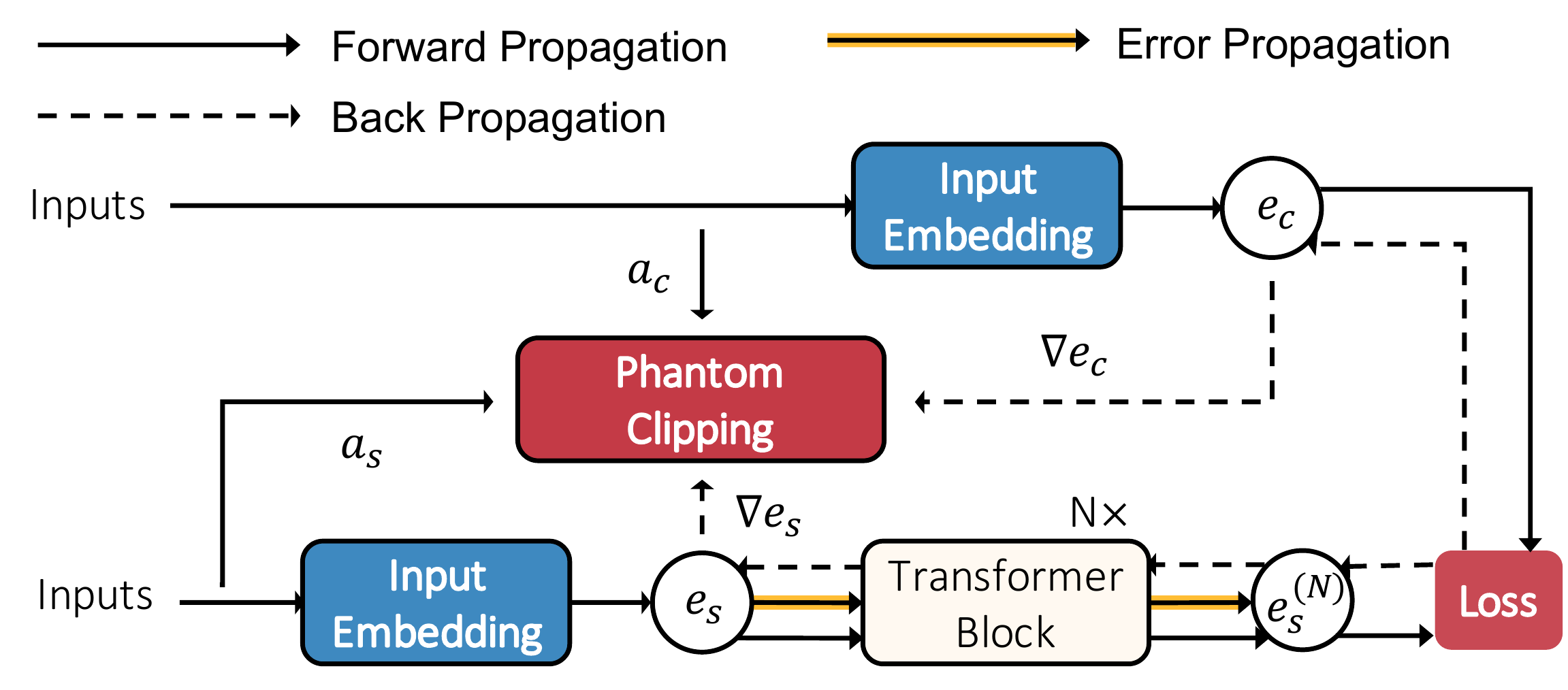}
\caption{Phantom Clipping}
\label{fig:pc}
\end{subfigure}
\begin{subfigure}{0.28\textwidth}
\centering
\includegraphics[width=3.8cm]{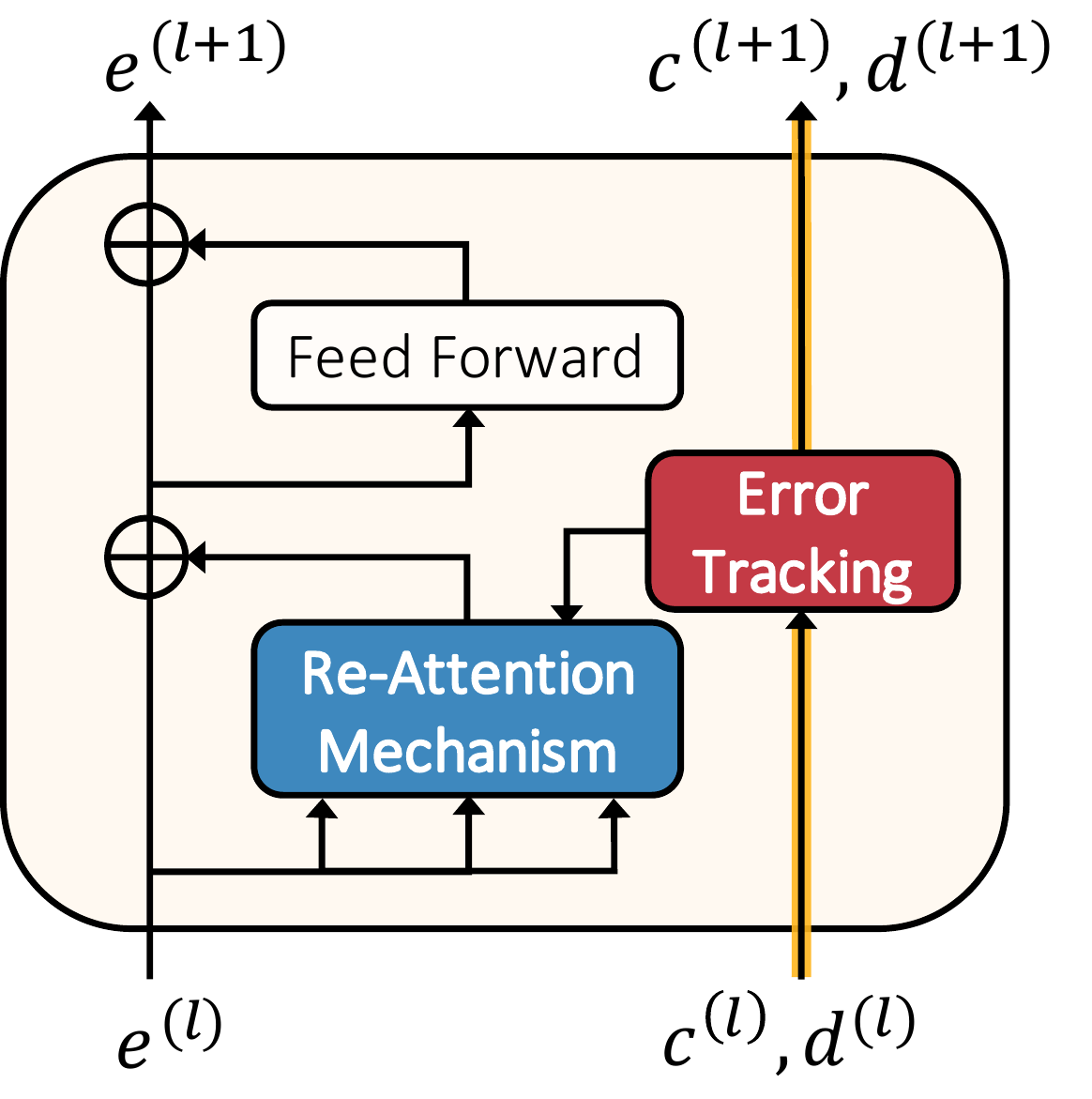}
\caption{Re-Attention Mechanism}
\label{fig:reattn}
\end{subfigure}
\caption{The DPFormer - methodology overview. \textbf{Left:} Phantom Clipping makes private training of Transformers almost as efficient as non-private training. \textbf{Right:} Re-Attention Mechanism aims to combat the attention distraction during private training and thereby improves the model performance.}
\label{fig:DPFormer}
\end{figure}

\section{Phantom Clipping}
\subsection{Motivation: The Indispensability of Parameter Sharing as a Form of Inductive Bias}
A recent advancement, Ghost Clipping~\cite{li2022large},  generalizes the Goodfellow trick~\cite{goodfellow2015efficient} to facilitate efficient clipping during the fine-tuning of Transformer-based language models on private data, without necessitating per-sample gradient computation.
Notwithstanding its considerable benefits in training Transformer models using DP-SGD as compared to other libraries or implementations (for instance, Opacus~\cite{yousefpour2021opacus}, JAX~\cite{subramani2021enabling}), Ghost Clipping presents a limitation in its lack of support for parameter sharing (i.e., the practice that ties the parameter of the input embedding and the output embedding layer together).
Although certain tasks, such as fine-tuning language models with differential privacy can achieve high accuracy without parameter sharing, our investigation reveal that when training  with DP-SGD, parameter sharing of the embedding layer becomes essential.

\begin{figure}[h]

\centering
\begin{subfigure}{0.3\textwidth}
\centering
\includegraphics[height=3.5cm]{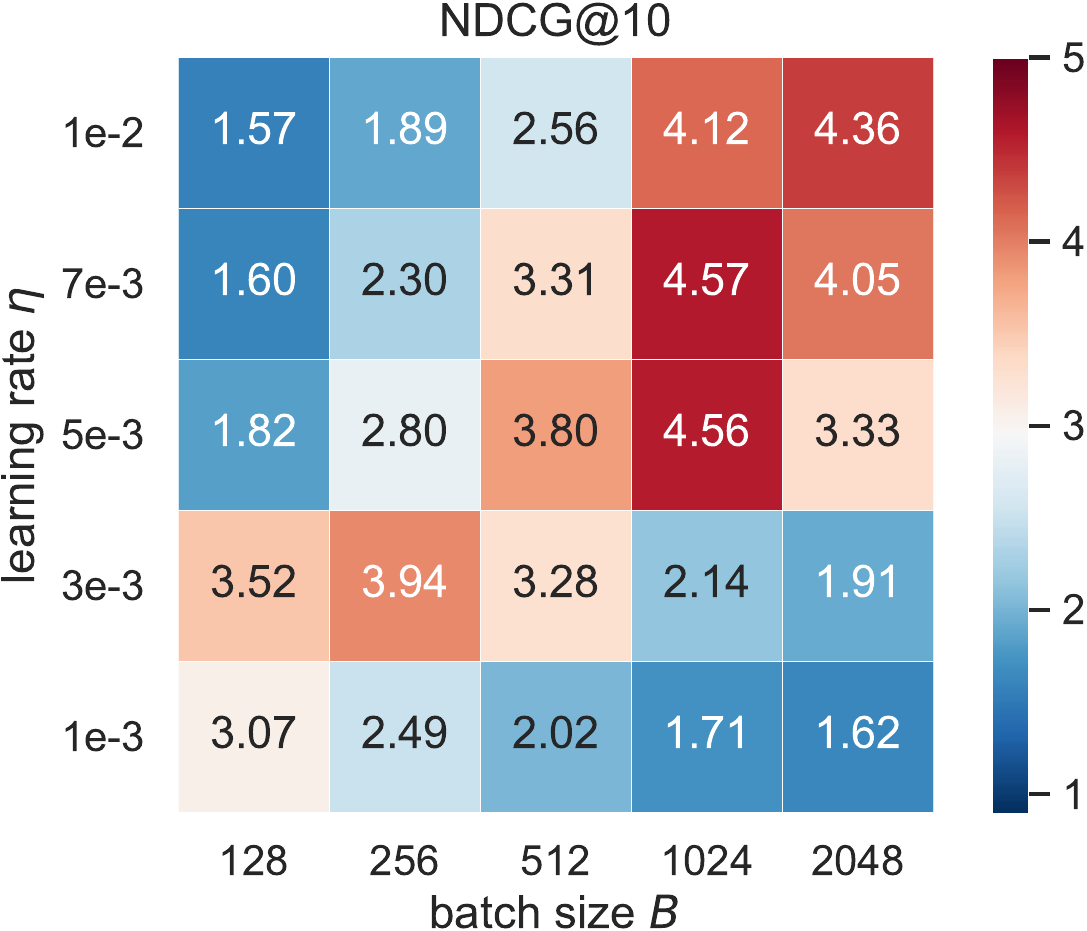}
\caption{parameter sharing}
\label{fig:heat1}
\end{subfigure}
\begin{subfigure}{0.3\textwidth}
\centering
\includegraphics[height=3.5cm]{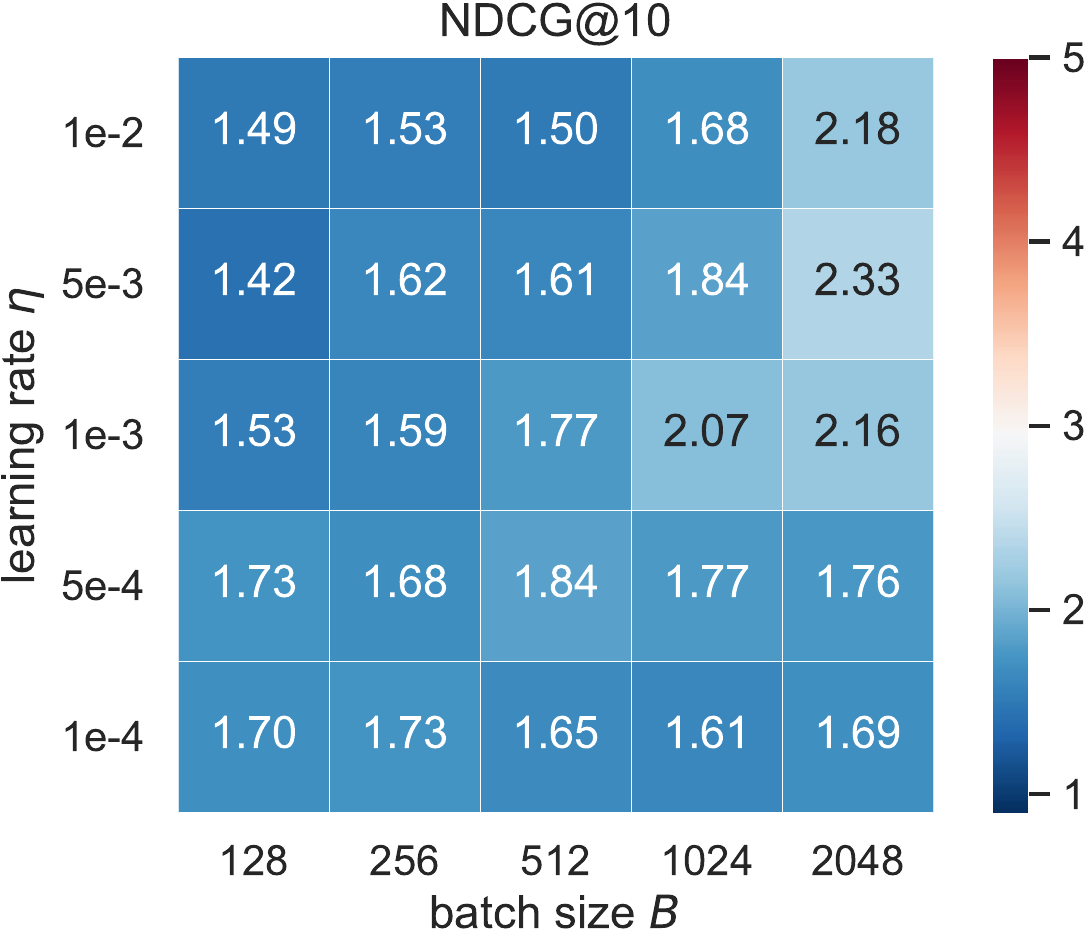}
\caption{w/o parameter sharing}
\label{fig:heat2}
\end{subfigure}
\begin{subfigure}{0.3\textwidth}
\centering
\includegraphics[height=3.5cm]{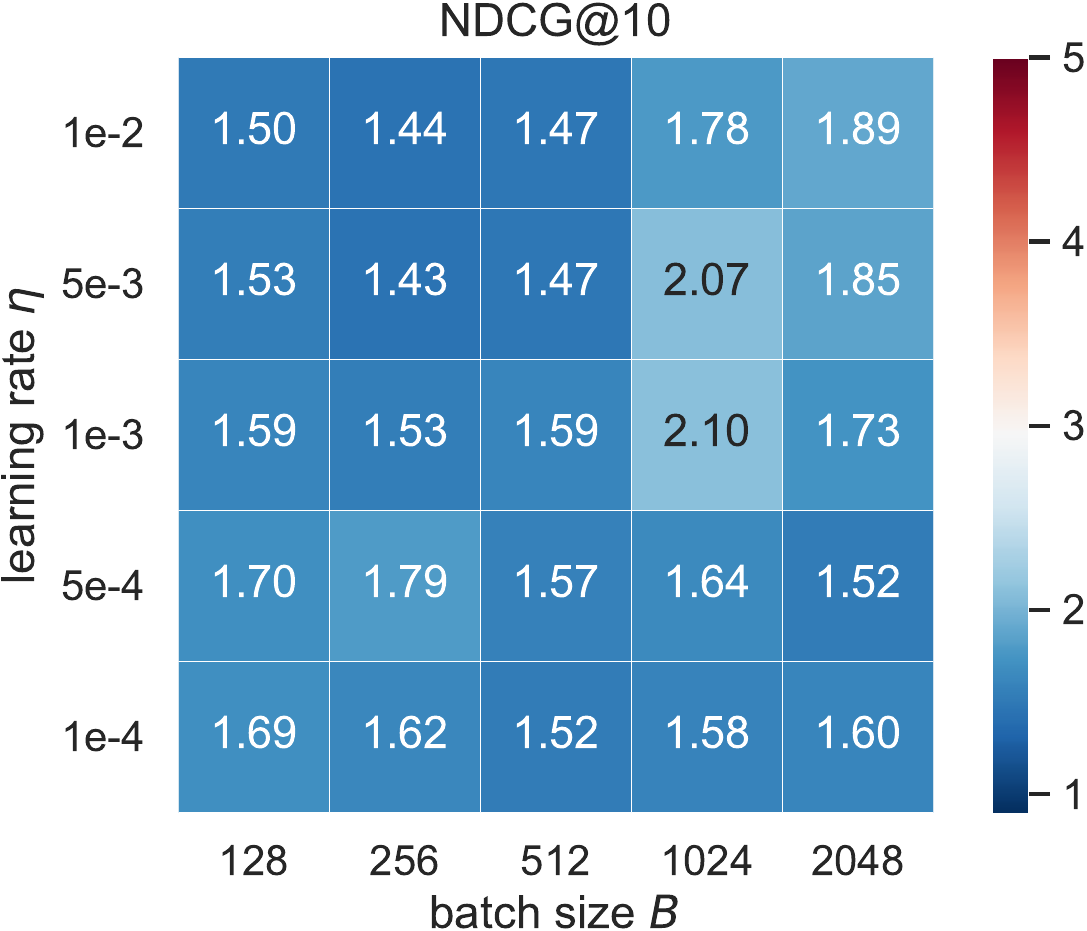}
\caption{halved dimension in (b)}
\label{fig:heat3}
\end{subfigure}
\caption{Numbers are NDCG(\%)@10 (higher is better) of the privately trained model (with $\varepsilon$ set to 5) on MovieLens (Figure~\ref{fig:dataset}). Parameter sharing for the embedding layer yields consistent and significant performance gains over the non-sharing setting in private training. The optimal hyperparameter configuration is always using a large batch size (with a large learning rate).}
\label{fig:embsharing}
\end{figure}

In more detail, to elucidate the role of parameter sharing when training with DP-SGD, we conduct experiments under the following three settings: (1) parameter sharing of the embedding layer, which aligns with the standard treatment in Transformer; (2) no parameter sharing; and (3) no parameter sharing coupled with a reduced embedding dimension by half.
Note that the third setting is included to account for the potential impact of model dimension on accuracy in private training, given the difference in the number of parameters between models with and without parameter sharing. Model performance across different hyperparameters is shown in Figure.~\ref{fig:embsharing}. 

The consistency and significance of the performance improvement brought by parameter sharing during private training are not hard to perceive. The essence of embedding sharing lies in the assumption that, by tying the embedding of the input and output layers, the representation of each token remains consistent throughout its retrieval. This inductive bias enhances the statistical efficiency of the model, enabling improved generalization. When training with DP-SGD on limited training data, the model must independently uncover this relationship from the noisy gradients with a low signal-to-noise ratio, heightening the convergence challenge.

\subsection{Efficient Per-Sample Gradient Norm Computation with Parameter Sharing}
In this section, we present \emph{Phantom Clipping}, a technique for efficient private training of Transformers without the need for instantiating per-sample gradient. Phantom Clipping is built upon Ghost Clipping, with additional support for efficient gradient clipping of the shared embedding layer.


\begin{figure}[h]
\vspace{-0.3cm}
\centering
\begin{subfigure}{0.48\textwidth}
\centering
\setlength{\abovecaptionskip}{0.cm}
\includegraphics[width=5cm]{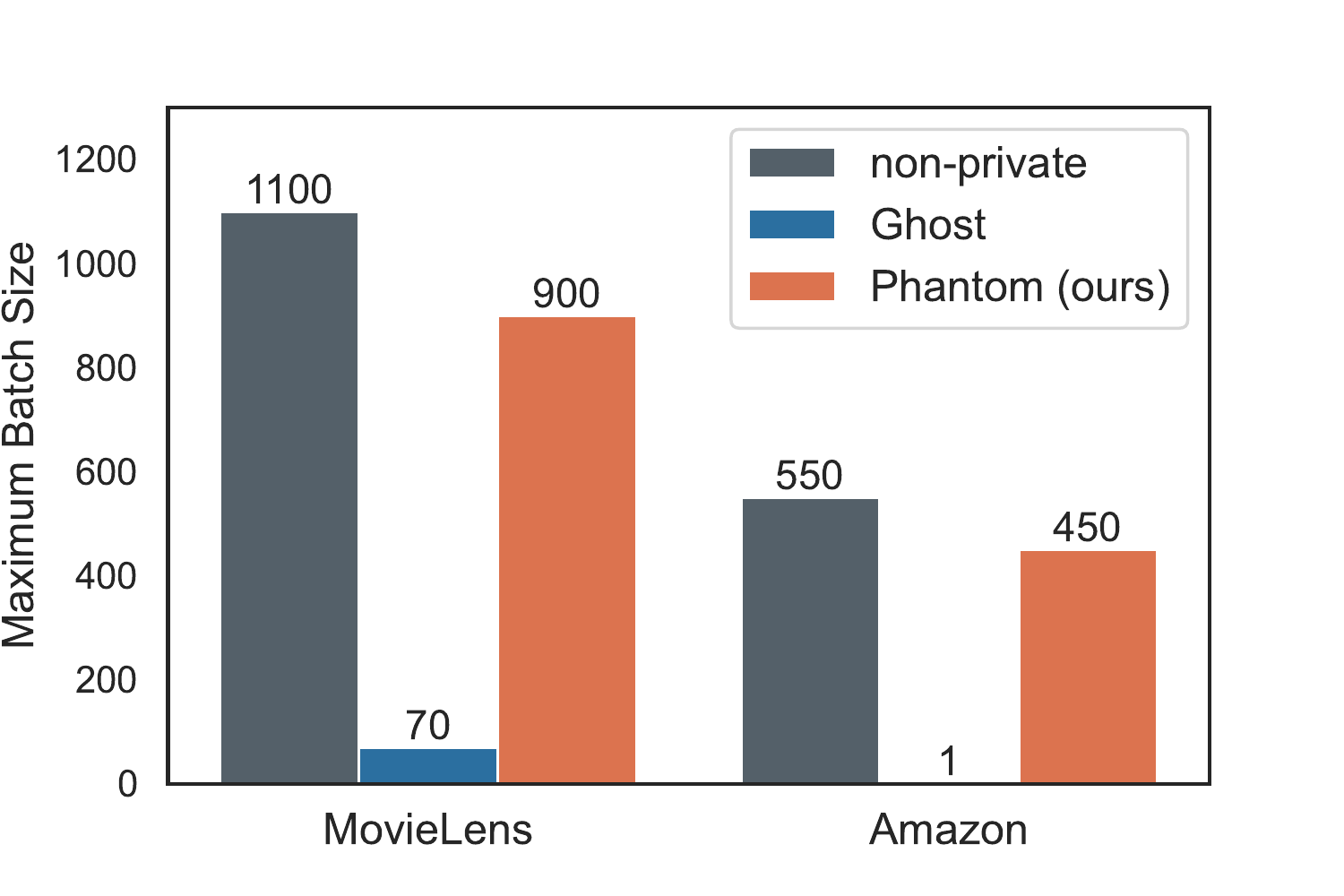}
\caption{Memory efficiency}
\label{fig:Memory}
\end{subfigure}
\begin{subfigure}{0.48\textwidth}
\centering
\setlength{\abovecaptionskip}{0.cm}
\includegraphics[width=5cm]{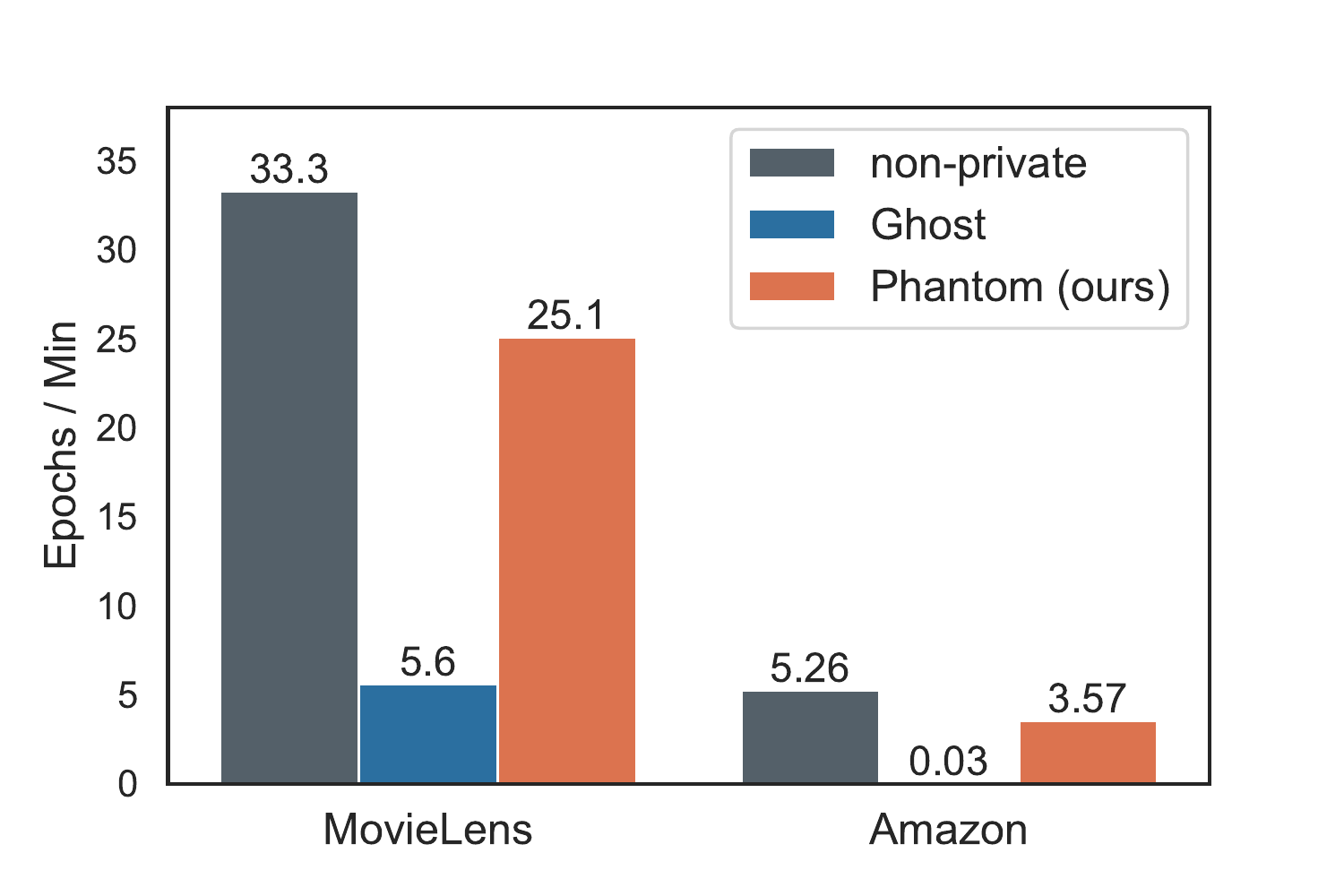}
\caption{Training speed}
\label{fig:Throughput}
\end{subfigure}

\label{fig:phantom}
\caption{\textbf{Left:} Phantom Clipping is 10-400$\times$ more memory efficient than Ghost Clipping and is almost as efficient as non-private training. \textbf{Right:} Phantom Clipping is 4-100$\times$ faster than Ghost Clipping, having comparable training speed with non-private training.}
\vspace{-0.2cm}
\end{figure}

Recall that the computational bottleneck of gradient clipping in Equation~(\ref{eq:DPSGD}) lies in the calculation of the per-sample gradient norm i.e., $\|g_{i}\|$. As the L2 norm of a vector can be decomposed cross arbitrary dimensions, for example, $\left\|(a, b, c)\right\| = \left\| (\|a\|, \|[b, c]\|)\right\|$. It suffices to consider the per-sample gradient norm $\|g_{i, E}\|$ of the embedding layer $E$ because the disparity due to parameter sharing lies solely in the shared embedding, and other layers can be handled akin to Ghost Clipping.
After obtaining the gradient norm via $\|g_{i}\|$ = $\|(\|g_{i, E}\|, \|g_{i, -E}\|)\|$, the next step is to scale the gradient by a factor of $\operatorname{Clip}_C(\|g_{i}\|)$ to bound its sensitivity. This can either be accomplished by re-scaling the loss $\mathcal{L}_i$ through this factor, followed by a second backpropagation~\cite{lee2021scaling}, or by manually scaling the gradient as demonstrated by~\cite{bu2022differentially}.

The challenge of evaluating $\|g_{i, E}\|$ efficiently without instantiating $g_{i, E}$ stems from the non-plain feed-forward (and symmetrically, backward propagation) topology caused by parameter sharing. See Figure~\ref{fig:pc} for a visual illustration, where the shared embedding leads to two branches of the backpropagation.
Despite this complexity, our Phantom Clip, described below, is capable of achieving this task. The derivation is deferred to Appendix~\ref{apdx:pfphan}. 

\begin{claim}
    (\textbf{Phantom Clipping}) Let $a_{i, s} \in \{0, 1\}^{L\times M}$ (or $a_{i, c} \in \{0, 1\}^{M\times M}$) be the one-hot encodings of the input sequence $s_i$ (or those of the candidate tokens for the output probability) in a minibatch. Let $e_{i, s} \in \mathbb{R}^{L\times d}$ (or $e_{i, c} \in \mathbb{R}^{M\times d}$) be output of the (shared) embedding layer $E$ when fed into $a_{i, s}$ (or $a_{i, c}$).
    Then the norm of the per-sample gradient with respect to $E$ can be efficiently evaluated as
\begin{equation}
    \|g_{i, E}\| = \left( \langle a_{i, s}a_{i, s}^T, \nabla e_{i, s} \nabla e_{i, s}^T \rangle^2 + \|\nabla e_{i, c}\|^2 + 2 \cdot \langle \nabla e_{i, s}, a_{i, s}^T \nabla e_{i, c} \rangle  \right)^{\frac{1}{2}},
    \label{eq:phantom}
\end{equation}
where $\nabla e_{i, s}:= \partial \mathcal{L}_i / \partial e_{i, s} \in \mathbb{R}^{L\times d}$, $\nabla e_{i, c} := \partial \mathcal{L}_i / \partial e_{i, c} \in \mathbb{R}^{M\times d}$, and $\langle\cdot, \cdot \rangle$ is the inner product of two matrices being of the same shape.
\end{claim}

\noindent \textbf{Memory Complexity} We study the additional memory footprint required by Phantom Clipping. Due to the storage of $a_sa_s^T$ (and $\nabla e_{s} \nabla e_{s}^T) \in \mathbb{R}^{B\times L\times L}$ in the first term of Equation~(\ref{eq:phantom}) (note that $a^T \nabla e_{c}$ is merely an indexing operation, requiring no additional memory), \emph{Phantom Clipping} has overhead memory complexity of $O(BL^2)$. As a comparison, Ghost Clipping has a memory complexity of $O(BT^2)$ when the input to the layer $a_i$ has the shape of $\mathbb{R}^{T\times \cdot}$. Hence its memory complexity for the two embedding layers is $O(BM^2 + BL^2)$ where $M$ is the vocabulary size. Typically, we have $L^2 \ll M$, that is, the length of the sentence is much smaller than the vocabulary size. Thus the memory overhead complexity of $\emph{Phantom Clipping}$ is only a negligible portion of that of $\emph{Ghost Clipping}$.

We implement our Phantom Clipping based on AWS's fastDP\footnote{\url{https://github.com/awslabs/fast-differential-privacy}} library, which has implemented Ghost Clipping.
We then empirically compare our $\emph{Phantom Clipping}$ with Ghost Clipping in terms of both memory footprint and training speed on real-world datasets\footnote{Since Ghost Clipping does not support parameter sharing, its results are obtained from training models without embedding sharing. This leads to more model parameters. For a fair comparison, we halve its embedding dimension to $d_E/2$, ending up with a similar number of parameters as in the model with embedding sharing.} (see Figure~\ref{fig:dataset} for details of the datasets).
Figure~\ref{fig:Memory} shows the maximum batch size that can fit into a  Tesla V100 GPU (16 GB of VRAM). It can be seen that our technique is much more memory friendly.
It allows up to $450\times$ larger batch size compared with Ghost Clipping on Amazon, almost as large as those in non-private training.
Figure~\ref{fig:Throughput} shows the training speed on a single Tesla V100 GPU. It allows up to $100\times$ training speedup in practice compared to Ghost Clipping, achieving  0.68$\times$ training speed of the non-private version.

\section{Re-Attention Mechanism}
\label{sec:reattn}
\begin{figure}[h]
\vspace{-0.5cm}
\centering
\setlength{\abovecaptionskip}{0.cm}
\includegraphics[width=14cm]{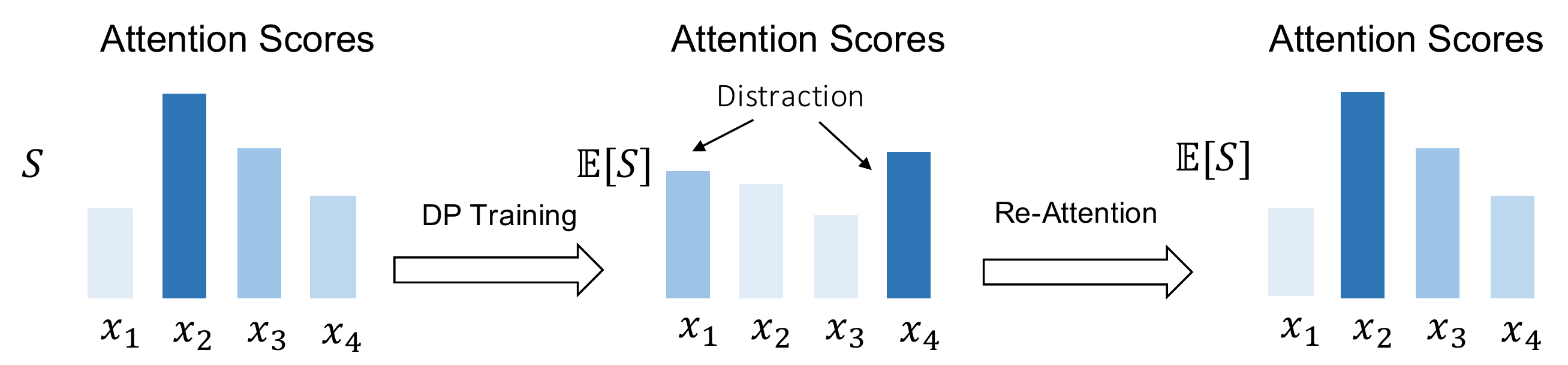}

\caption{Illustration of the proposed Re-Attention Mechanism. Fix some query $q$, consider the attention scores with respect to $x_1$, $x_2$, $x_3$ and $x_4$, where $x_1$ and $x_4$ are assumed to be tail tokens. \textbf{Left:} The attention scores in non-private setting, i.e., ground truth, with the highest attention on tokens $x_2$ and $x_3$. \textbf{Middle:} Expectation of attention scores in DP training, where the attention is distracted to $x_4$ and $x_1$ due to the relatively higher uncertainty level of $x_1$ and $x_4$. \textbf{Right:} Re-Attention mechanism is designed to handle attention distraction by correcting the attention scores.
}
\label{fig:reattnscore}
\vspace{-0.2cm}
\end{figure}
\subsection{Motivation: The Attention Distraction Phenomenon}

Recall that the Attention Mechanism, as the key component of the Transformer, given a query, calculates attention scores pertaining to tokens of relevance. Our key observation is that, over the randomness of the attention keys for the tokens of interest\footnote{In unambiguous contexts, `token variance' will denote the variance of the relevant representation associated with that token, like its attention key, $K_i$, or its embedding, $E_i$.}, the expectation of the attention scores will be distorted, particularly, mindlessly leaning towards tokens with high variance, regardless of their actual relevance. 
Refer to Figure~\ref{fig:reattnscore} for a visual illustration of this concept of attention distraction.

To shed light on this phenomenon, we offer a theoretical analysis.
Let us fix some query $q$ and denote the attention key of token $i$ as $K_i$. Since DP-SGD injects Gaussian noise into the model, it is natural to assume $K_i$  follows a Gaussian distribution with mean $k_i$ and variance $\sigma^2_i$. 
We denote $S_i$ as the random variable of the attention score assigned to token $i$. With some basic algebraic manipulation and applying the theory of extreme value~\cite{coles2001introduction}, we can recast the formula for attention scores as follows\footnote{For ease of notation, we omit the constant factor (i.e., $1/\sqrt{d}$) in attention computation.},
\begin{equation}
\begin{split}
    S_i = \frac{\exp{\langle q, K_i\rangle}}{\sum_{j=1}^L \exp{\langle q, K_j\rangle}} 
    &= \exp{\left(\langle q, K_i\rangle - \log \sum_{j=1}^{L} \exp{\langle q, K_j\rangle}\right)} \\
    &= \exp{\left(\langle q, K_i\rangle - \E_{\gamma} [\operatorname{max}_j\{\langle q, K_j\rangle + \gamma \}]\right)}, \\
\end{split}
\label{eq:logsumexp}
\end{equation}
where $\gamma$ is distributed as a standard Gumbel. 
Let us consider some token $i'$ that should have attracted little attention given the query $q$,
then the expectation of the noisy maximum $\E_{\gamma} [\operatorname{max}_j\{\langle q, K_j\rangle + \gamma \}]$ can be approximated by $\operatorname{max}_{j\neq i'}{\langle q, K_j\rangle} + \zeta$, where $\zeta = \E [\gamma] = \pi^2/6$.
Taking the expectation of Equation~(\ref{eq:logsumexp}) over $K_{i'}$, and leveraging the fact $\E [\exp(X)] = \exp(\E[X])\exp(\operatorname{Var}[X]/2)$ when $X$ follows a Gaussian distribution, we then arrive at the following conclusion
\begin{equation}
\begin{split}
    \E_{K_{i'}} [S_{i'}] 
    &\approx  \E_{K_{i'}} [\exp{(\langle q, K_{i'}\rangle - (\operatorname{max}_{j\neq i'}\{\langle q, K_j\rangle\}+ \zeta))}]\\
    &= \underbrace{\exp{(\langle q, k_{i'}\rangle - \widetilde{M})}}_{\rm attentive~relevance} \cdot \underbrace{\exp{\left(C\sigma^2_{i'}/2\right)}}_{\rm multiplicative~error},
\end{split}
\label{eq:reattnbias}
\end{equation}
where $\widetilde{M}=(\operatorname{max}_{j\neq i'}\{\langle q, K_j\rangle\}+ \zeta)$ and the last equality leverages the fact that $\langle q, K_{i'} \rangle \sim \mathcal{N}(\langle q, k_{i'}\rangle, C\sigma^2)$ with $C=\langle q, q\rangle$.
As a result, tokens with higher variance result in inflated attention scores due to increased multiplicative bias, distracting attention from more deserving tokens, given that token $i'$ is presupposed to garner little attention under query $q$. If all tokens have similar variance or variance terms are negligible, the negative effects of this attention diversion are reduced. However, in less ideal conditions, especially with long-tailed data, this attention distraction could hinder the Transformer's training, thereby degrading model utility.




\subsection{Re-Attention via Error Tracking }
At its core, the Re-Attention Mechanism is designed to mitigate attention distraction during the learning process via debiasing the attention scores. 
To achieve this, it is natural to track the variance term identified as the error multiplier in Equation~(\ref{eq:reattnbias}). In the following discussion, we elaborate on the methodology employed for tracking this error term during private training.

\subsubsection{Error Instantiation}
Let us focus on the source of the randomness which leads to the attention distraction phenomenon, that is, the DP noise injected into the model gradients.
Inspired by~\cite{li2022large}, we propose the idea of \emph{effective error}, which is a probabilistic treatment of the \emph{effective noise multiplier} in~\cite{li2022large}, proposed for model with sequential input. Effective error is used as an estimate of the uncertainty level underlying the model parameters, where the randomness is over DP noise.
\begin{definition}
\textbf{Effective Error:} The effective error $\sigma_{\eff}^{\theta}$ associated with the model parameter $\theta$ is defined as
\begin{equation}
\begin{split}
    \sigma_{\eff}^{\theta} = \frac{\sigma_{\dip}}{B_{\eff}^{\theta}}, ~~~~~\text{where} ~B_{\eff}^{\theta} = \E_{\mathcal{B}\stackrel{\tiny\text{i.i.d}}{~\sim~}\mathcal{D}^B} \left[\sum_{i=1}^B \mathbb{I}\left[ R_{\theta}(\mathcal{B}_i)\right]\right],
\end{split}
\label{eq:effectiveerror}
\end{equation}
where $B\in \mathbb{N}$ is the batch size, $\mathcal{B}\in \mathbb{N}^{B\times L}$ is the minibatch, i.i.d. sampled from training data distribution $\mathcal{D}$ (note that $\mathcal{B}_i$ is a sequence of tokens), $\sigma_{\dip}$ is the DP noise multiplier in Equation~(\ref{eq:DPSGD}), and $\mathbb{I\left(\cdot\right)}$ is the indicator function, $R_\theta(\cdot) = 1$ if $\mathcal{B}_i$ has relevance with $\theta$, for example, $R_{E_i}(\mathcal{B}_j) = \mathbb{I}[{\rm token}~i\in \mathcal{B}_j]$ where $E_i$ is the embedding for token $i$, and $R_{W}(\mathcal{B}_j) = 1$ where $W$ is the parameter within Transformer Block (see Figure~\ref{fig:DPFormer}).
\begin{remark}
    \emph{Effective error} recovers \emph{effective noise multiplier} when the model has no embedding layer, for example, an MLP model. In that case, $\sigma_{\eff}^{\theta}=\sigma_{\dip}/B$.
\end{remark}

\end{definition}
We then have the following claims for obtaining effective error of the Transformer's parameters. See Appendix~\ref{apdx:errorinst} for detailed derivation.
\begin{claim}
    For each layer parameterized by $W$ within the Transformer block, its effective error is $\sigma_{\eff}^{W} = \sigma_{\dip} / B$.
\label{cl:errorinst1}
\end{claim}
\begin{claim}
    For the embedding layer $E$, effective error of token $i$ is $\sigma_{\eff}^{E_i} = \sigma_{\dip} / (B\cdot p_i)$, where $p_i$ is the frequency of token $i$ (i.e., the probability of token $i$'s occurrence in data).
\label{cl:errorinst2}
\end{claim}

\subsubsection{Error Propagation}

Given the effective errors of the embedding layer and of the Transformer encoder, our goal is to obtain the error term $\sigma_i$ identified in Equation~(\ref{eq:reattnbias}) for each attention computation. 
Notably, this issue of error tracking aligns with studies in Bayesian deep learning~\cite{wang2020survey}, a field primarily focused on quantifying prediction uncertainty to enhance the robustness and reliability of machine learning systems.
While our primary interest lies in unbiased attention score computation during private training, we can leverage and adapt existing methodologies in Bayesian deep learning to achieve this distinct goal. 
Specifically, given the input embedding along with its effective error,  we propagate the effective error through Transformer layers (see Figure~\ref{fig:DPFormer}), with the goal of obtaining $\sigma_i$ for each attention calculation. 
We denote the output of the $l$-th layer by the random variable $X^{(l)}$.
Given the output distribution $X^{(l-1)}$ of the preceding layer, the distribution $X^{(l)}$ can be computed layer-by-layer as follows, 
\begin{equation}
\begin{split}
    p(X^{(l)}|X^{(0)}) = \E_{X^{(l-1)}|X^{(0)}} \left[p\left(X^{(l)} | X^{(l-1)}\right)\right] = \E_{X^{(l-1)}|X^{(0)}} \left[p^d\left(X^{(l)}_i | X^{(l-1)}\right)\right],
\end{split}
\end{equation}
where the last equality is due to the isometric Gaussian noise of DP (see Equation~\ref{eq:DPSGD}), i.e., each dimension is independently and identically distributed. Based on Variational Inference~\cite{kingma2013auto}, we can use an approximating distribution $q$ to approximate the computationally intractable distribution $p$, where $q(X^{(l)})$ follows a Gaussian distribution of mean $\mu$ and variance $\sigma^2$. Note that minimizing the KL divergence of $KL(p(X^{(l)}|X^{(0)}) || q(X^{(l)}))$ reduces to matching the moments of $q(X^{(l)})$ to $p(X^{(l)}|X^{(0)})$. Since the mean and variance\footnote{Note that for a Gaussian distribution, (i) mean and variance, (ii) the first two moments, and (iii) natural parameter, are equivalent in the sense of mutual convertibility. We will use them interchangeably.} are sufficient statistics for Gaussian distribution, propagating the distribution reduces to propagating its natural parameters~\cite{wang2016natural}. For linear layers coupled with a coordinate-wise non-linear activation, the statistics can be computed by analytic expressions using existing techniques from Probabilistic Neural Networks~\cite{wang2016natural,shekhovtsov2019feed,gast2018lightweight,postels2019sampling,morales-alvarez2021activationlevel}. 
Concretely, for linear transformation, $X^{(l)}=X^{(l-1)}W$, we can propagate the variance as
\begin{equation}
    \sigma_{X^{(l)}}^2 = \sigma_{X^{(l)}}^2 \cdot \sigma_W^2 + \sigma_{X^{(l)}}^2 \cdot \mu_W^2 +  \sigma_W^2 \cdot (\mu_{X^{(l)}})^2
    \label{eq:linearnpn}.
\end{equation}
For nonlinear activation functions, e.g., $X^{(l)}=\operatorname{ReLU}\left(X^{(l-1)}\right)$, we can propagate the variance as
\begin{equation}
\begin{split}
    \sigma_{X^{(l)}}^2 = \Phi(\frac{c}{\sqrt{d}})(c^2+d) + \frac{c\sqrt{d}}{\sqrt{2\pi}}\exp(-\frac{1}{2}\frac{c^2}{d}) - c^2,
\end{split}    
\label{eq:relunpn}
\end{equation}
where $\Phi(\cdot)$ is the cumulative density function (CDF) of the standard Gaussian distribution, $c$ and $d$ are the natural parameter of  $X^{(l-1)}$. For completeness, derivation is included  in Appendix~\ref{apdx:npnproof}.

All in all, we can obtain the output distribution of layer $(l)$ via analytic expression  in terms of the natural parameter~\cite{wang2016natural} of the preceding layer's output distribution as 
\begin{equation}
(c^{(l)}, d^{(l)}) = \mathcal{F}(c^{(l-1)}, d^{(l-1)}),~~~\sigma^2 = \mathcal{T}(c^{(l)}, d^{(l)}).
\label{eq:nonprop}
\end{equation}

Nevertheless, a nuanced difference exists between our error propagation and existing techniques encapsulated in Equation~(\ref{eq:nonprop}). In the Bayesian approach, the model parameter is directly associated with the mean $\mu_W$ in Equation~(\ref{eq:linearnpn}). During private training, however, we can only access the noisy parameter after the injection of DP noise. Interestingly, access to this noisy parameter can be interpreted as a single sampling opportunity from its underlying Gaussian distribution, which can then be viewed as a one-time Markov Chain sampling~\cite{wang2015privacy}. Therefore, the noisy parameter can serve as an estimate of its mean.
In addition, unlike variance propagation in Bayesian deep learning, the error propagation here incurs minimal computational and memory overhead as the effective error can be represented in scalar (again, due to the isometric DP noise), plus the propagation is performed via analytical expressions.

\textbf{Re-Attention.} With the effective error tracked, we then proceed to mitigate the attention distraction identified in Equation~(\ref{eq:reattnbias}) via $S_i \leftarrow S_i / \exp \left[C\sigma^2_i/2\right]$, obtaining unbiased attention scores.

In summary, we can propagate and track the effective error through the layers: given the natural parameter of $X^{(l-1)}$, the variance can be estimated using analytic expressions, which then can be used to correct the attention scores. This methodology of error instantiation from DP-SGD multiplier, propagation through layers, and attention debiasing forms the crux of our Re-Attention Mechanism. 


\section{Experiments}

\begin{wrapfigure}{r}{7.5cm}
\vspace{-0.7cm}
    \setlength{\abovecaptionskip}{0.cm}
    \centering
    \includegraphics[width=0.99\linewidth]{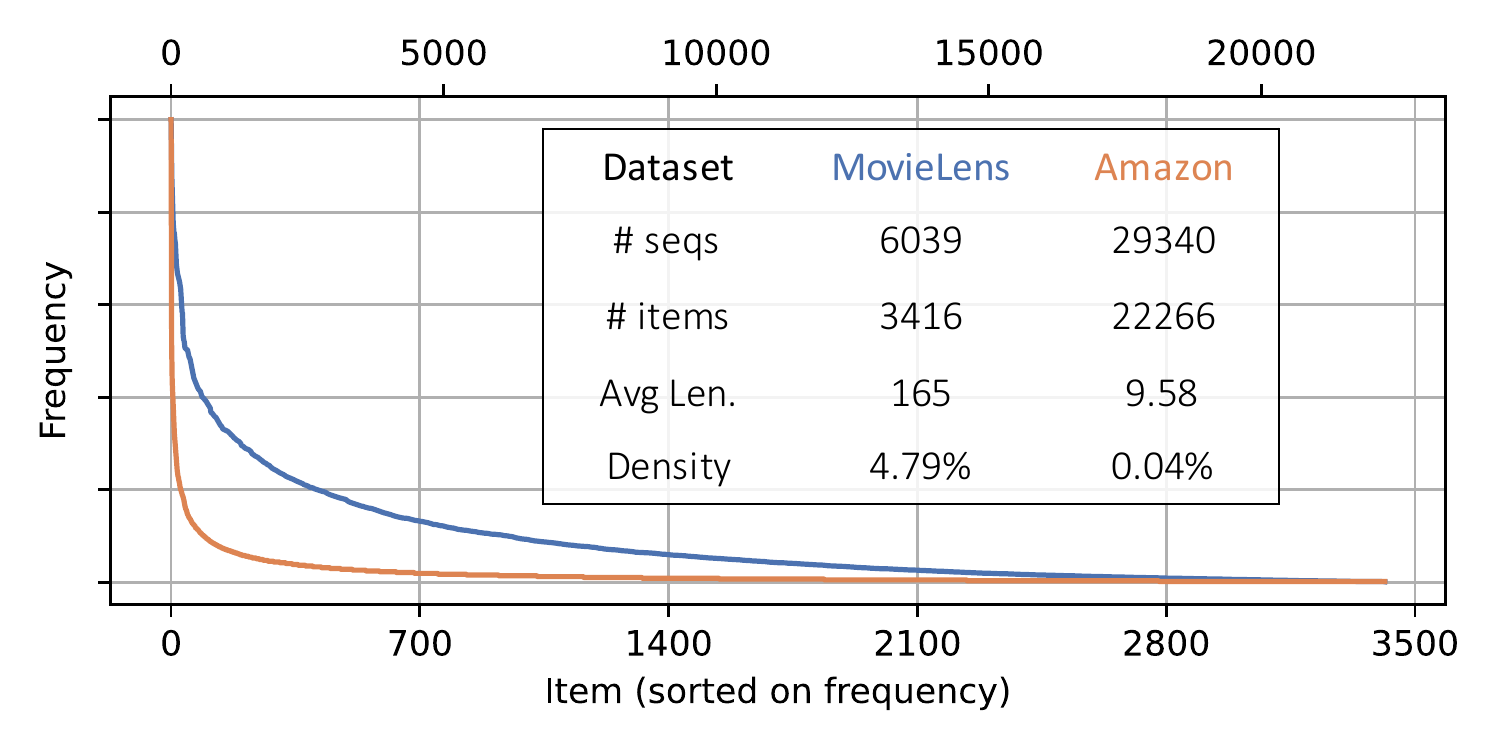}
    \caption{Data in the real-world scenarios exhibits long-tailed (also known as, power-law) distribution.}
    \label{fig:dataset}
\vspace{-0.3cm}
\end{wrapfigure}

\textbf{Datasets and Prevalence of Long-Tailed Distributions.} 
We conduct experiments on two public datasets collected from real-world scenarios: MovieLens~\cite{harper2015movielens} and Amazon~\cite{mcauley2015image}. Figure~\ref{fig:dataset} shows their data distributions, illustrating the prevalence of long-tailed distributions, where a small number of items are extremely popular and have relatively high frequency while other items occur infrequently. The embedded table above the `long tail' reports the statistics of the two datasets, showing that the two datasets vary significantly in size and sparsity.
More details on datasets can be found in Appendix~\ref{apdx:exp:dataset}.

\textbf{Baselines and Implementation Details.} 
We compare our DPFormer with vanilla Transformer~\cite{vaswani2017attention} (i.e., the one without Re-Attention Mechanism), vanilla Transformer without parameter sharing, GRU~\cite{cho2014learning}, and LSTM~\cite{hochreiter1997long}. For a fair comparison, embedding sharing is applied for all evaluated methods if not explicitly stated.
The number of epochs is set to 100, where the first 20\% of epochs are used for learning rate warm-up. After that, we linearly decay the learning rate through the remaining epochs.
Following~\cite{bu2022automatic,yang2022normalized}, we normalize the gradients and set the clipping norm $C$ to 1, which eliminates the hyperparameter tuning for clipping norm $C$. The model dimension is set to 64. For privacy accounting, we fix the total training epochs (iterations) and derive the noise required for each iteration from the preset privacy budget $\varepsilon$. We consider $\varepsilon=5, 8, 10$.


\begin{table}[h]
\scriptsize
\centering
\vspace{-0.5cm}
\caption{Best results (\%) on MovieLens at different privacy levels.}
\begin{tabular}{@{}clcccccc@{}}
\toprule
\multicolumn{2}{c}{DP Guarantee} & \multicolumn{2}{c}{$\varepsilon=5$}   & \multicolumn{2}{c}{$\varepsilon=8$}   & \multicolumn{2}{c}{$\varepsilon=10$} \\ \midrule
\multicolumn{2}{c}{Metric}       & NDCG@10 & \multicolumn{1}{c|}{HIT@10} & NDCG@10 & \multicolumn{1}{c|}{HIT@10} & NDCG@10           & HIT@10           \\ \midrule\midrule
\multicolumn{2}{c}{\textsc{GRU}}  & 2.26 $\pm$ 0.04 & \multicolumn{1}{c|}{4.58 $\pm$ 0.09} & 2.40  $\pm$ 0.03   & \multicolumn{1}{c|}{4.75 $\pm$ 0.20} & 2.81 $\pm$ 0.03 & 5.53 $\pm$ 0.05 \\
\multicolumn{2}{c}{\textsc{LSTM}}   & 2.65 $\pm$ 0.07 & \multicolumn{1}{c|}{5.08 $\pm$ 0.08} & 2.76 $\pm$ 0.03 & \multicolumn{1}{c|}{5.41 $\pm$ 0.06} & 2.95 $\pm$ 0.03 & 5.55 $\pm$ 0.06 \\
\multicolumn{2}{c}{\textsc{Transformer w/o PS}}  & 2.33 $\pm$ 0.05 & \multicolumn{1}{c|}{4.47 $\pm$ 0.07} & 2.56 $\pm$ 0.03    & \multicolumn{1}{c|}{5.11 $\pm$ 0.05} & 2.74 $\pm$ 0.04  & 5.39 $\pm$ 0.08  \\
\multicolumn{2}{c}{\textsc{Transformer (Vanilla)}}  & \textbf{4.57 $\pm$ 0.26} & \multicolumn{1}{c|}{\textbf{8.69 $\pm$ 0.53}} & \textbf{7.05 $\pm$ 0.23}    & \multicolumn{1}{c|}{\textbf{13.17 $\pm$ 0.37}} & \textbf{7.99 $\pm$ 0.21} & \textbf{14.82 $\pm$ 0.38} \\ \midrule \midrule
\multicolumn{2}{c}{\textsc{DPFormer (Ours)}} & \textbf{5.88 $\pm$ 0.24}  & \multicolumn{1}{c|}{\textbf{11.13 $\pm$ 0.43}}     &  \textbf{7.70 $\pm$ 0.26} & \multicolumn{1}{c|}{\textbf{14.31 $\pm$ 0.37}}     & \textbf{8.42 $\pm$ 0.22} & \textbf{15.40 $\pm$ 0.32} \\ \midrule
\multicolumn{2}{c}{Relative Improvement} & \textcolor{darkgreen}{\textbf{29\%$\uparrow$}} &  \multicolumn{1}{c|}{\textcolor{darkgreen}{\textbf{28\%$\uparrow$}}}   & \textcolor{darkgreen}{\textbf{9.2\%$\uparrow$}} &  \multicolumn{1}{c|}{\textcolor{darkgreen}{\textbf{8.7\%$\uparrow$}}}     & \textcolor{darkgreen}{\textbf{5.4\%$\uparrow$}} & \textcolor{darkgreen}{\textbf{3.9\%$\uparrow$}} \\
\bottomrule
\end{tabular}
\label{tb:accml}
\end{table}

\begin{table}[h]
\scriptsize
\centering
\vspace{-0.2cm}
\caption{Best results (\%) on Amazon at different privacy levels.}
\begin{tabular}{@{}clcccccc@{}}
\toprule
\multicolumn{2}{c}{DP Guarantee} & \multicolumn{2}{c}{$\varepsilon=5$}   & \multicolumn{2}{c}{$\varepsilon=8$}   & \multicolumn{2}{c}{$\varepsilon=10$} \\ \midrule
\multicolumn{2}{c}{Metric}       & NDCG@10 & \multicolumn{1}{c|}{HIT@10} & NDCG@10 & \multicolumn{1}{c|}{HIT@10} & NDCG@10           & HIT@10           \\ \midrule\midrule
\multicolumn{2}{c}{\textsc{GRU}}  & 1.13 $\pm$ 0.02 & \multicolumn{1}{c|}{2.46 $\pm$ 0.03} & 1.33 $\pm$ 0.02    & \multicolumn{1}{c|}{2.22 $\pm$ 0.02} & 1.47 $\pm$ 0.03 & 2.48 $\pm$ 0.02 \\

\multicolumn{2}{c}{\textsc{LSTM}}   & 1.19 $\pm$ 0.01  & \multicolumn{1}{c|}{2.46 $\pm$ 0.04} & 1.23 $\pm$ 0.01  & \multicolumn{1}{c|}{2.46 $\pm$ 0.04} &  1.34 $\pm$ 0.01 &  2.51 $\pm$ 0.02 \\

\multicolumn{2}{c}{\textsc{Transformer w/o PS}}  & 1.16 $\pm$ 0.01 & \multicolumn{1}{c|}{2.36 $\pm$ 0.01} & 1.20 $\pm$ 0.02    & \multicolumn{1}{c|}{2.38 $\pm$ 0.01} & 1.40 $\pm$ 0.01  & 2.47 $\pm$ 0.02  \\

\multicolumn{2}{c}{\textsc{Transformer (Vanilla)}}  & \textbf{1.37 $\pm$ 0.04} & \multicolumn{1}{c|}{\textbf{2.47 $\pm$ 0.10}} & \textbf{1.54 $\pm$ 0.03}    & \multicolumn{1}{c|}{\textbf{2.77 $\pm$ 0.07}} & \textbf{1.57 $\pm$ 0.03}  & \textbf{2.83 $\pm$ 0.08}  \\ \midrule \midrule

\multicolumn{2}{c}{\textsc{DPFormer (Ours)}} & \textbf{1.64 $\pm$ 0.01}  & \multicolumn{1}{c|}{\textbf{3.01 $\pm$ 0.01}}     & \textbf{1.98 $\pm$ 0.05} & \multicolumn{1}{c|}{\textbf{3.70 $\pm$ 0.15}}     & \textbf{1.99 $\pm$ 0.04} & \textbf{3.73 $\pm$ 0.11} \\ \midrule

\multicolumn{2}{c}{Relative Improvement} & \textcolor{shilv}{\textbf{20\%$\uparrow$}} &  \multicolumn{1}{c|}{\textcolor{darkgreen}{\textbf{22\%$\uparrow$}}}   & \textcolor{shilv}{\textbf{28\%$\uparrow$}} &  \multicolumn{1}{c|}{\textcolor{darkgreen}{\textbf{34\%$\uparrow$}}}     & \textcolor{darkgreen}{\textbf{27\%$\uparrow$}} &\textcolor{darkgreen}{\textbf{31\%$\uparrow$}} \\
\bottomrule
\end{tabular}
\label{tb:accam}
\end{table}



\begin{figure}[ht!]
\centering
\begin{subfigure}{0.24\textwidth}
\centering
\setlength{\abovecaptionskip}{0.cm}
\includegraphics[width=3cm]{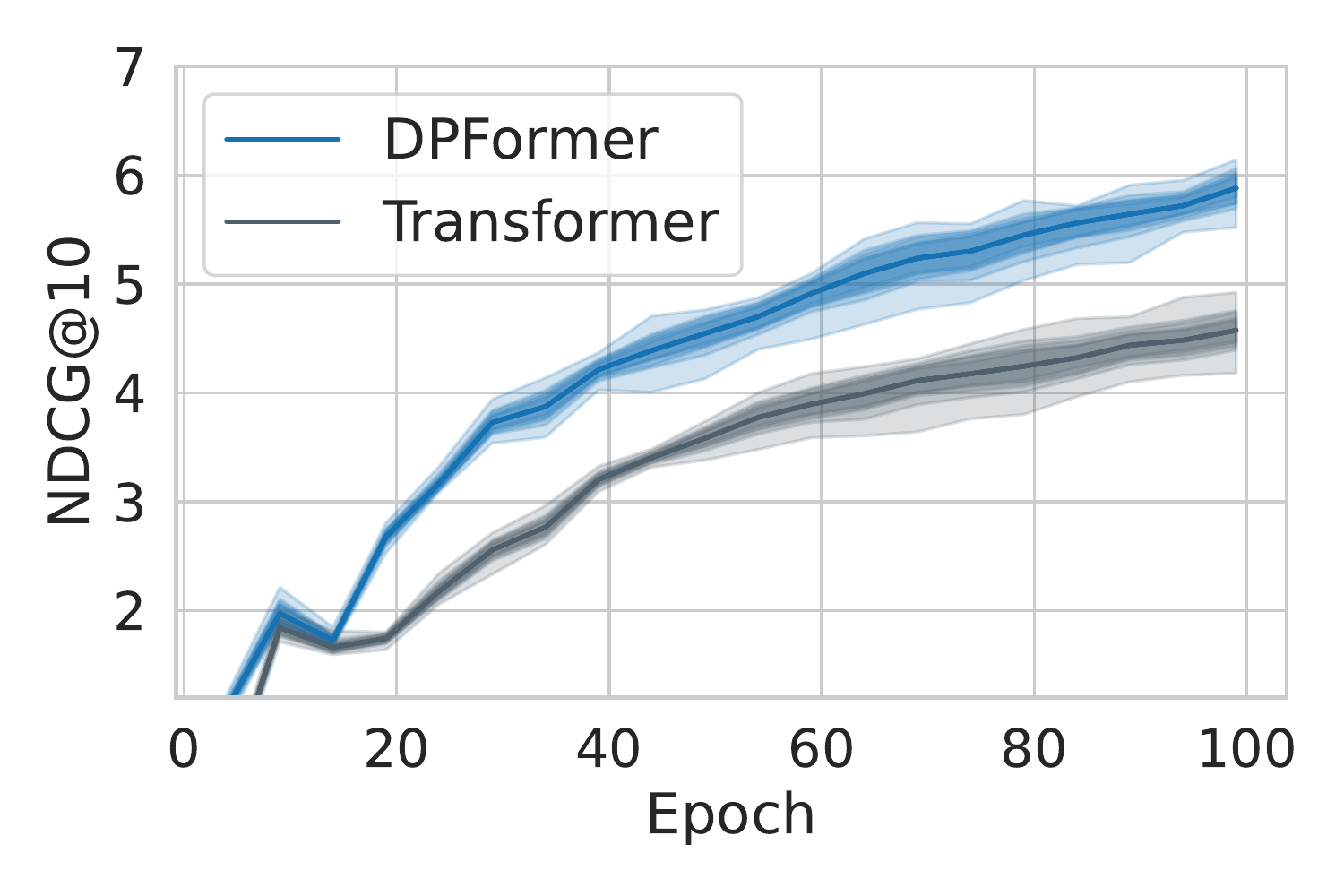}
\caption{MovieLens ($\varepsilon=5$)}
\label{fig:ndcg_amazon}
\end{subfigure}
\begin{subfigure}{0.24\textwidth}
\centering
\setlength{\abovecaptionskip}{0.cm}
\includegraphics[width=3cm]{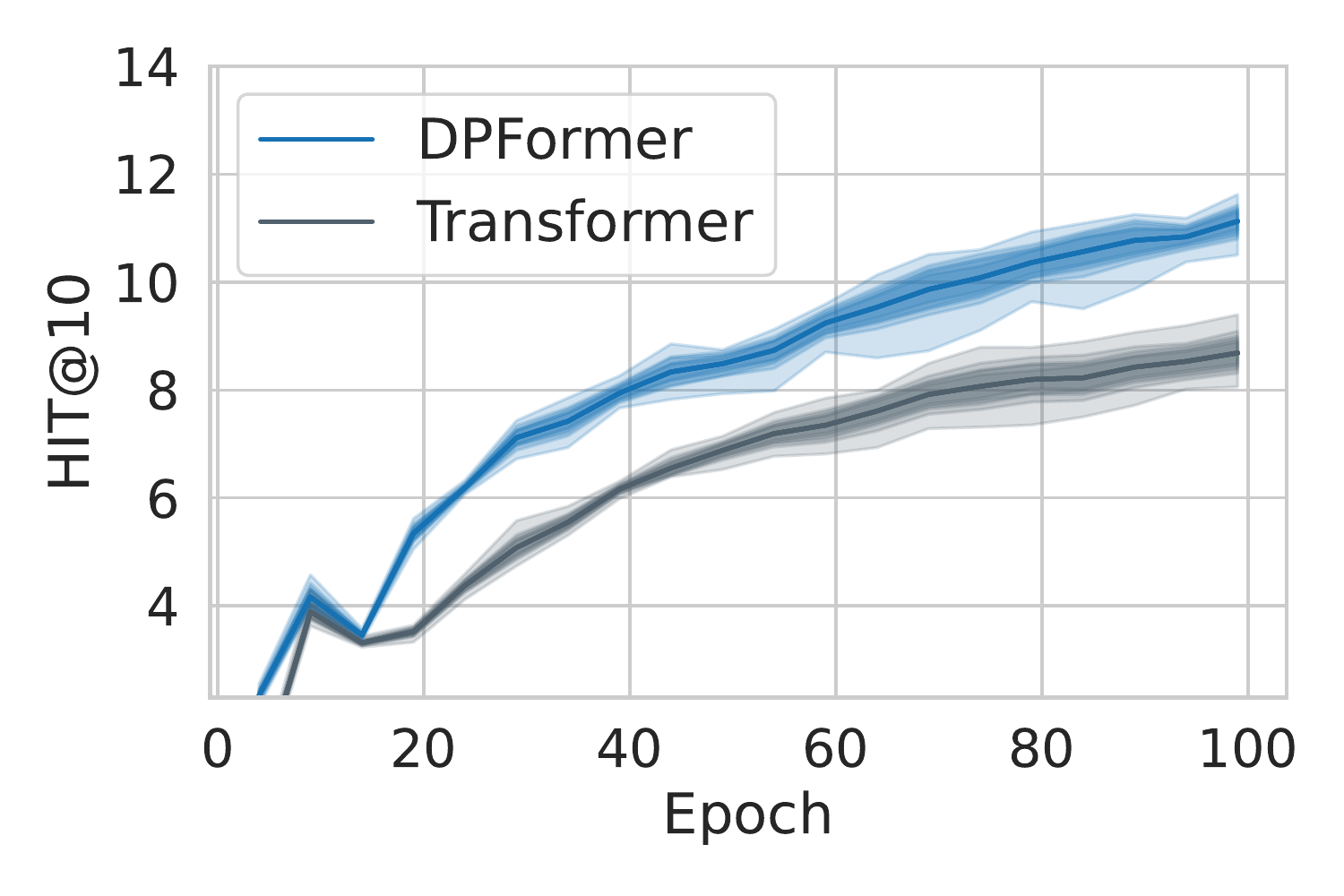}
\caption{MovieLens ($\varepsilon=5$)}
\label{fig:hit_ml1m_5}
\end{subfigure}
\begin{subfigure}{0.24\textwidth}
\centering
\setlength{\abovecaptionskip}{0.cm}
\includegraphics[width=3cm]{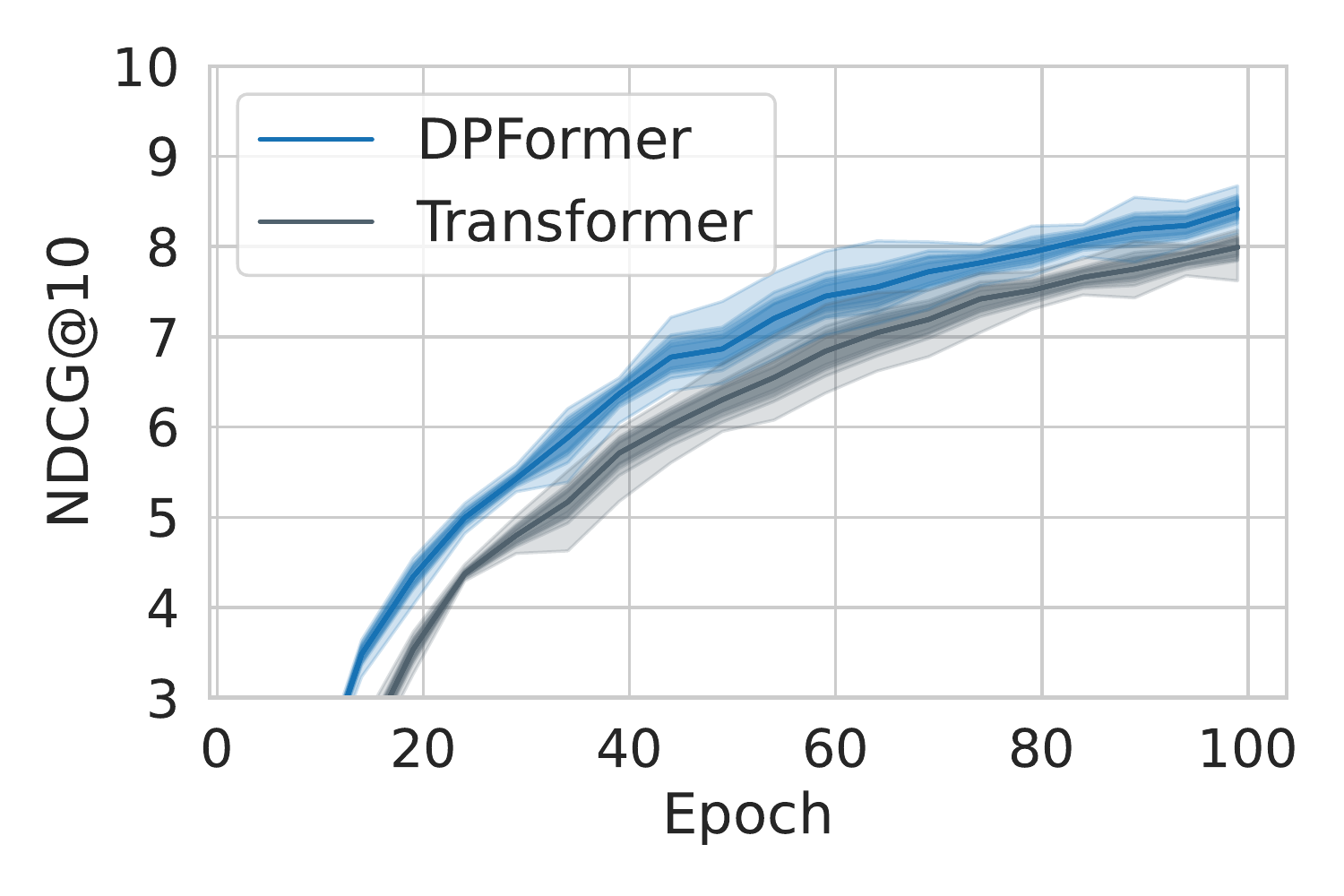}
\caption{MovieLens ($\varepsilon=10$)}
\label{fig:ndcg_ml1m_10}
\end{subfigure}
\begin{subfigure}{0.24\textwidth}
\centering
\setlength{\abovecaptionskip}{0.cm}
\includegraphics[width=3cm]{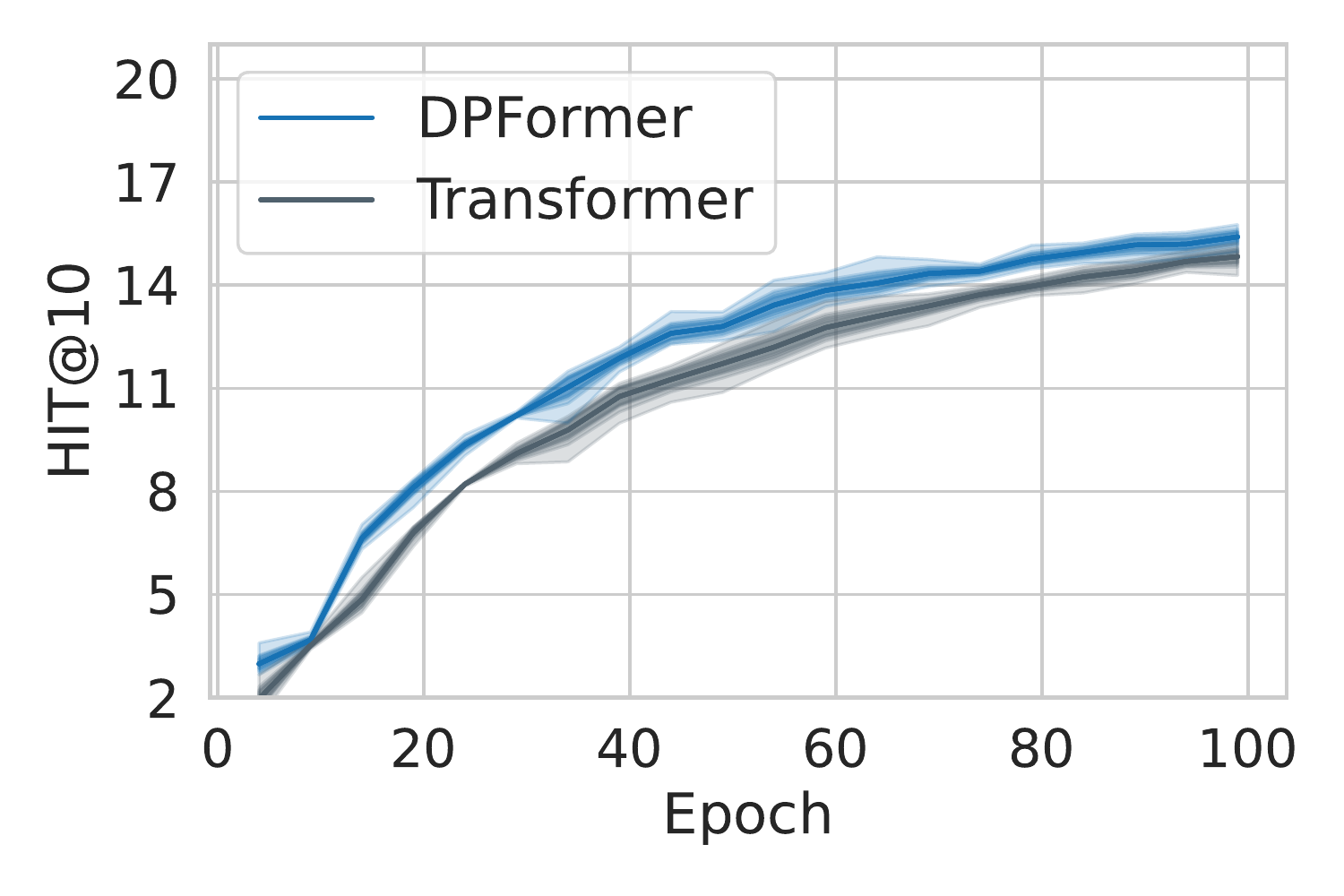}
\caption{MovieLens ($\varepsilon=10$)}
\label{fig:hit_ml1m_10}
\end{subfigure}

\begin{subfigure}{0.24\textwidth}
\centering
\setlength{\abovecaptionskip}{0.cm}
\includegraphics[width=3cm]{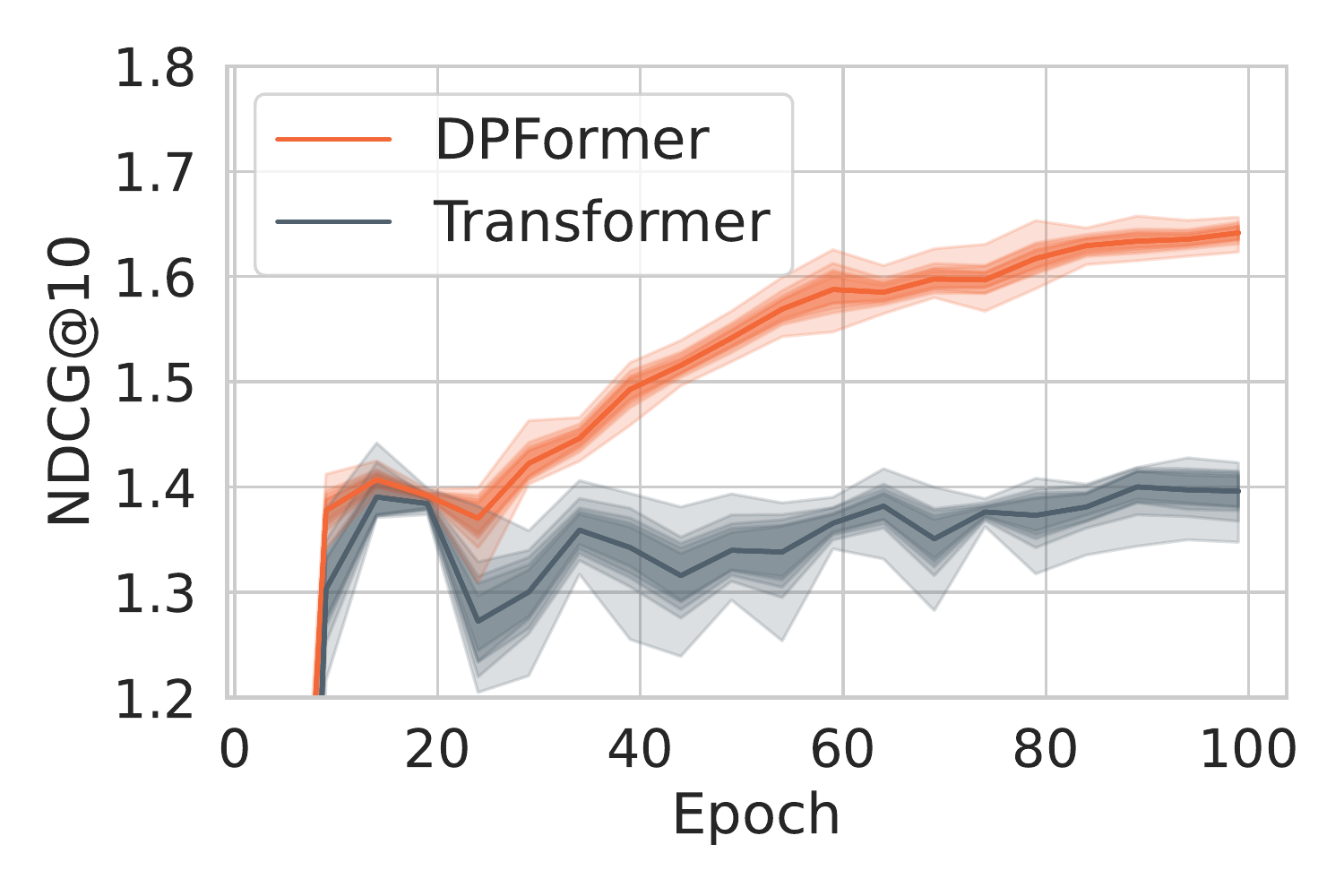}
\caption{Amazon ($\varepsilon=5$)}
\label{fig:ndcg_amazon_5}
\end{subfigure}
\begin{subfigure}{0.24\textwidth}
\centering
\setlength{\abovecaptionskip}{0.cm}
\includegraphics[width=3cm]{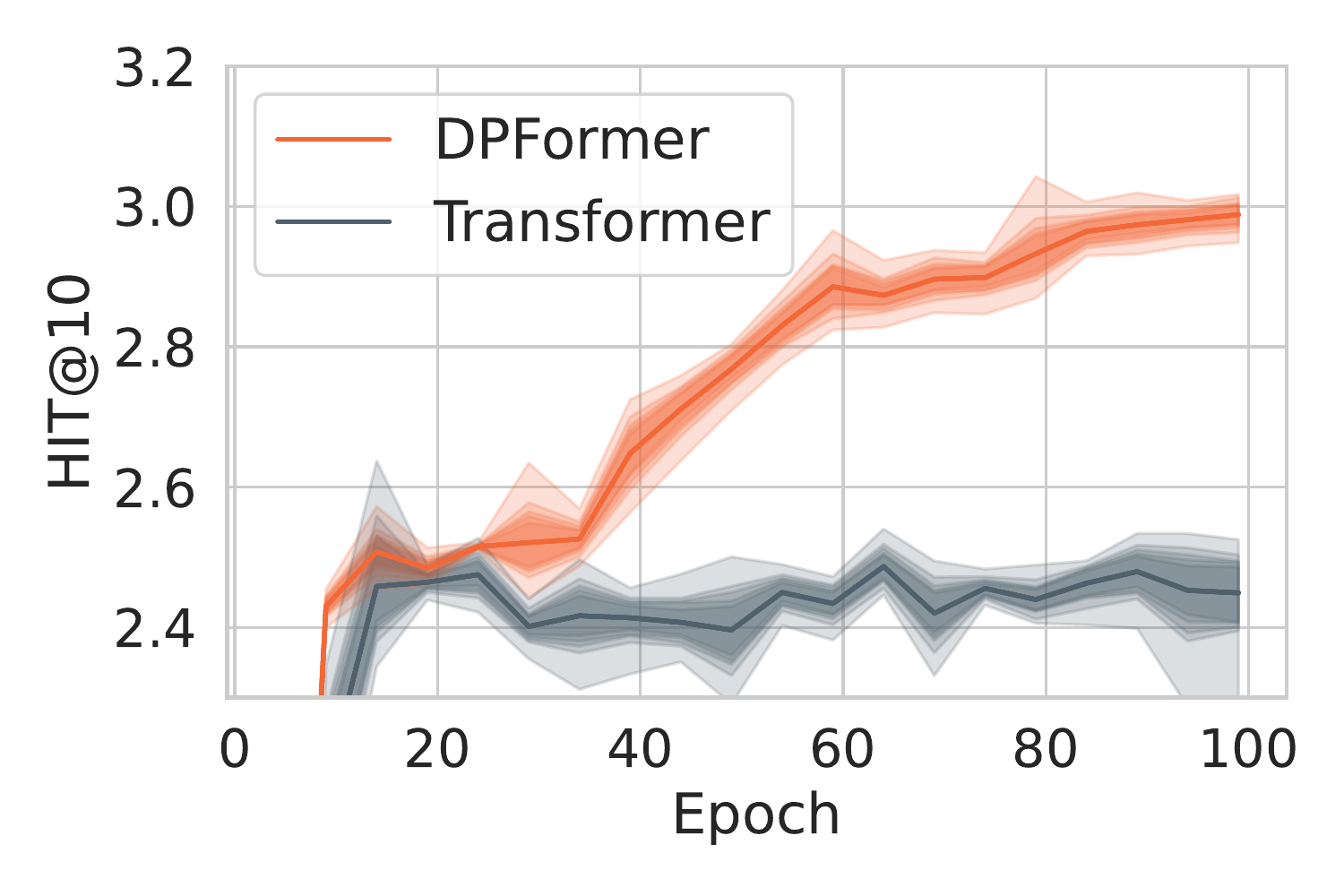}
\caption{Amazon ($\varepsilon=5$)}
\label{fig:hit_amazon_5}
\end{subfigure}
\begin{subfigure}{0.24\textwidth}
\centering
\setlength{\abovecaptionskip}{0.cm}
\includegraphics[width=3cm]{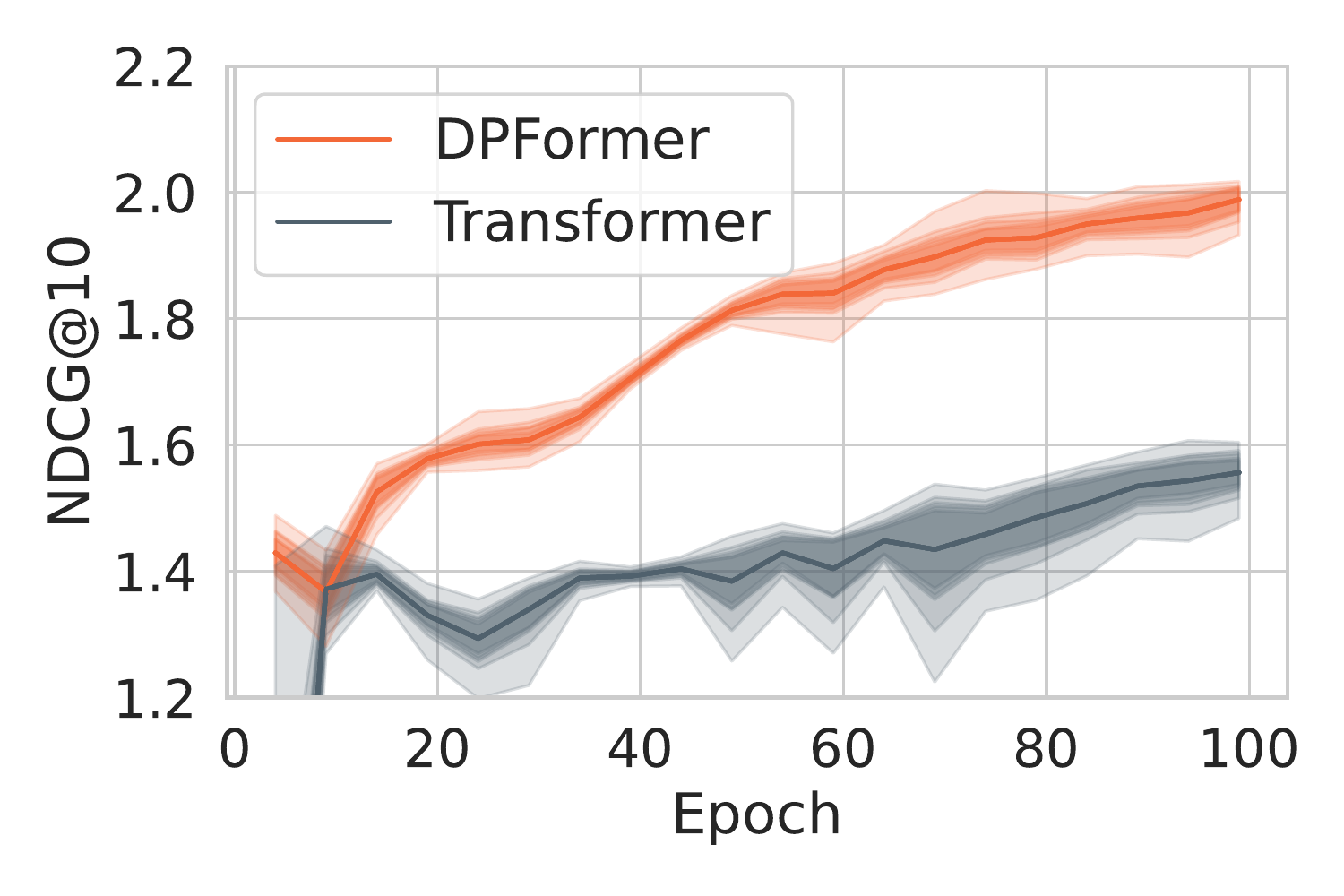}
\caption{Amazon ($\varepsilon=10$)}
\label{fig:ndcg_amazon_10}
\end{subfigure}
\begin{subfigure}{0.24\textwidth}
\centering
\setlength{\abovecaptionskip}{0.cm}
\includegraphics[width=3cm]{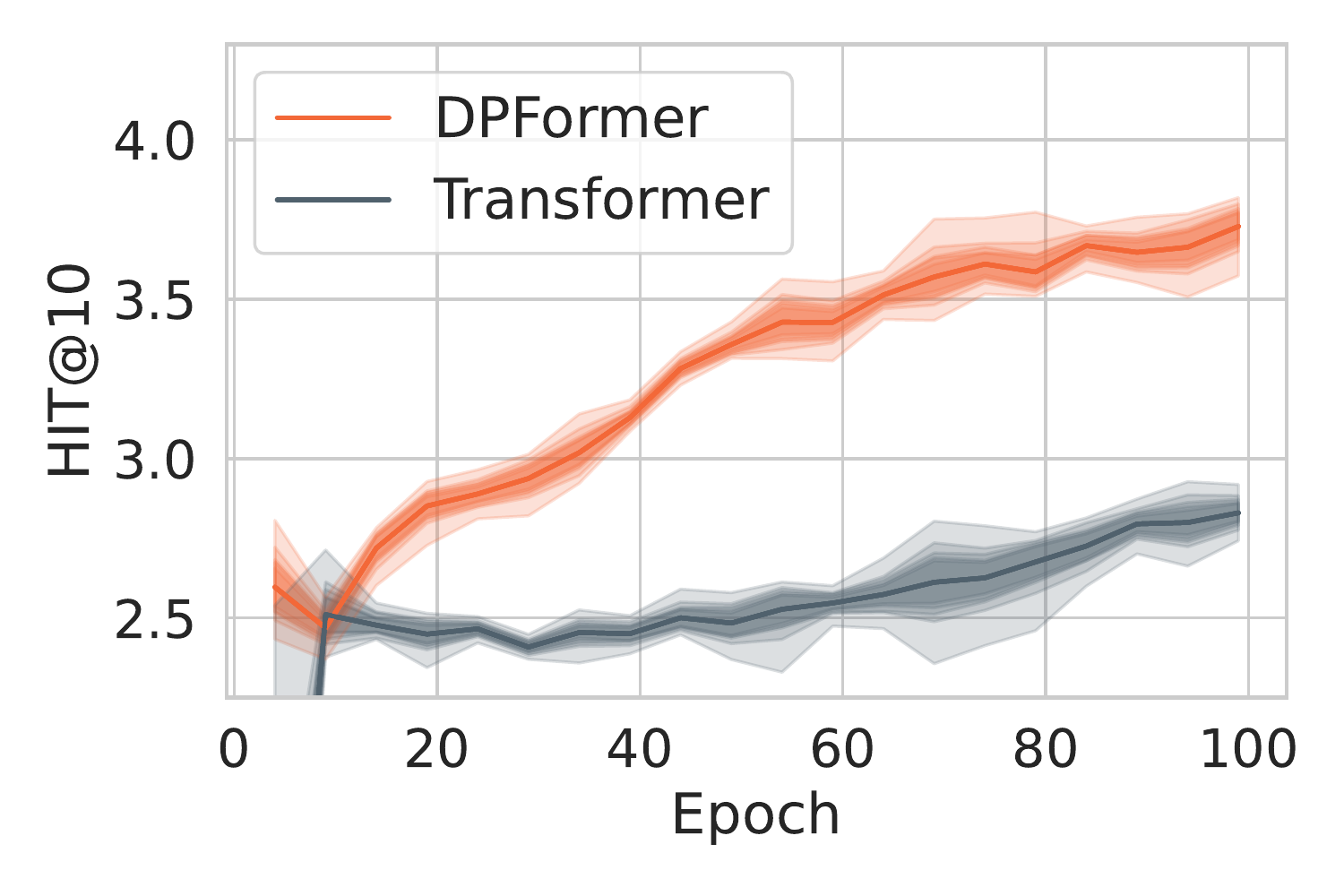}
\caption{Amazon ($\varepsilon=10$)}
\label{fig:hit_amazon_10}
\end{subfigure}

\caption{The Re-Attention Mechanism renders DPFormer notably more stable during private training. Each run is repeated five times with independent random seeds, with test accuracy (i.e., NDCG@10(\%) and HIT@10(\%)) reported every five epochs. The graduated shading (best viewed zoomed in) represents confidence intervals from 60\% to 100\%. }
\label{fig:convergence}
\end{figure}

Table~\ref{tb:accml} and Table~\ref{tb:accam} show the best\footnote{Strictly speaking, the process of hyperparameter tuning would cost privacy budget~\cite{papernot2022hyperparameter}, but is mainly of theoretical interest. We perform grid search on learning rate $\in\{10^{-3}, 3\times10^{-3}, 5\times10^{-3},  7\times10^{-3},  9\times10^{-3}\}$ and batch size $\in\{256, 512, 1024, 2048, 4096\}$ for each method, ensuring fair comparison.} NDCG@10 and HIT@10 for all the methods on MovieLens and Amazon. 
The vanilla Transformer outperforms all other baselines, reaffirming its dominance in sequential data modeling due to the Attention Mechanism. Our DPFormer, incorporating the Re-Attention Mechanism, further boosts the performance by around 20\% on average. Notably, under a low privacy budget ($\varepsilon=3$), DPFormer achieves a relative improvement of around 25\%, demonstrating its efficacy in attenuating attention distraction during private training. On MovieLens, as expected, the performance gain increases with decreasing privacy budget $\varepsilon$, i.e., increasing noise strength during training; this is because larger noise corresponds to more severe attention distraction, which better highlights the Re-Attention Mechanism's advantage. 
However, on Amazon, DPFormer achieves a smaller relative improvement at $\varepsilon=5$ than at $\varepsilon=10$. We suspect that this is due to the two datasets' differences in terms of sparsity (i.e., $1-$density in Figure~\ref{fig:dataset}) as well as the inherent hardness of training Transformer~\cite{zhang2019improving,xu2020optimizing,huang2020improving} and substantial DP noise.

Figure~\ref{fig:convergence} shows the model accuracy every five epochs during training. Evidently, the training dynamics of the vanilla Transformer, impacted by attention distraction, can suffer from high variance and/or substantial fluctuation, especially on Amazon. In contrast, DPFormer enjoys faster and smoother convergence, highlighting its superior training stability under differential privacy.

\begin{figure}[h]

\centering
\begin{subfigure}{0.24\textwidth}
\centering
\includegraphics[height=2.7cm]{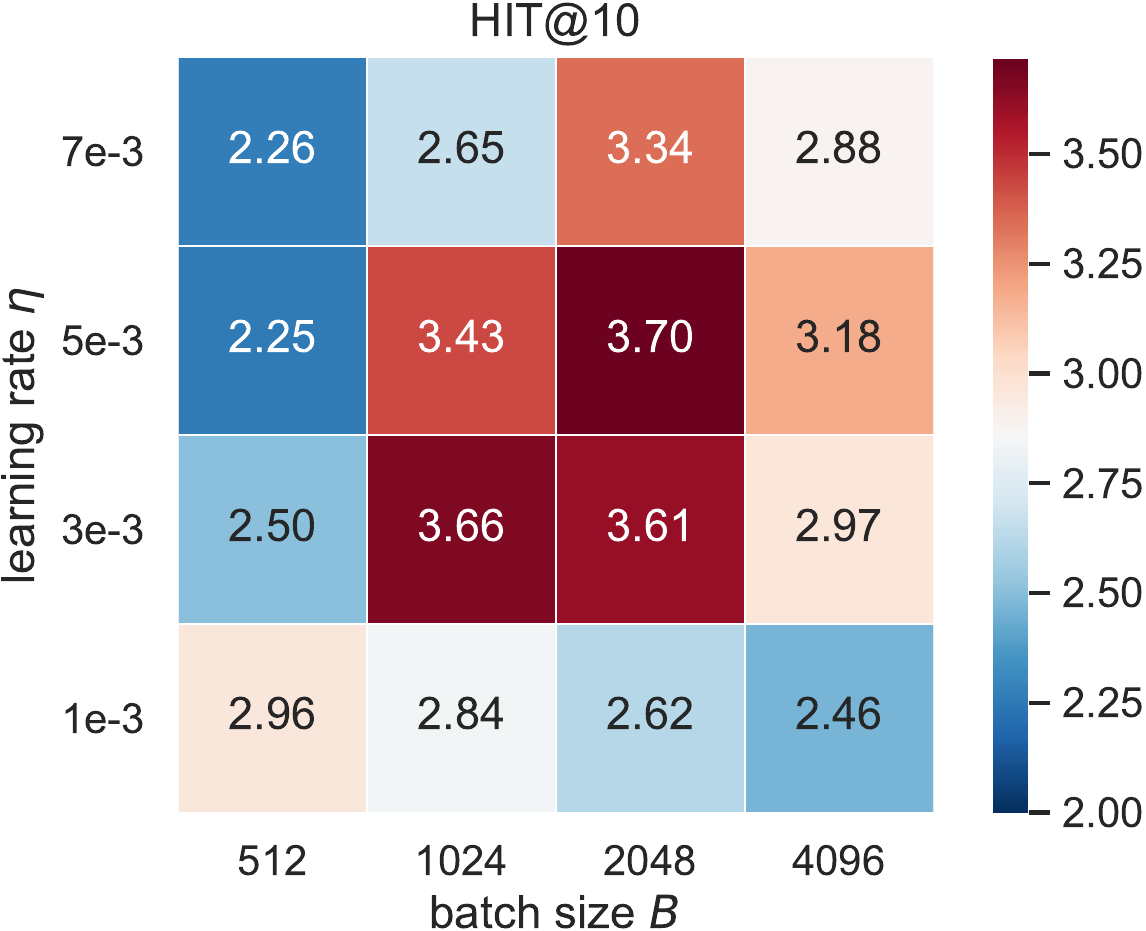}
\caption{DPFormer}
\label{fig:grad_ndcg_ama_dp}
\end{subfigure}
\begin{subfigure}{0.24\textwidth}
\centering
\includegraphics[height=2.7cm]{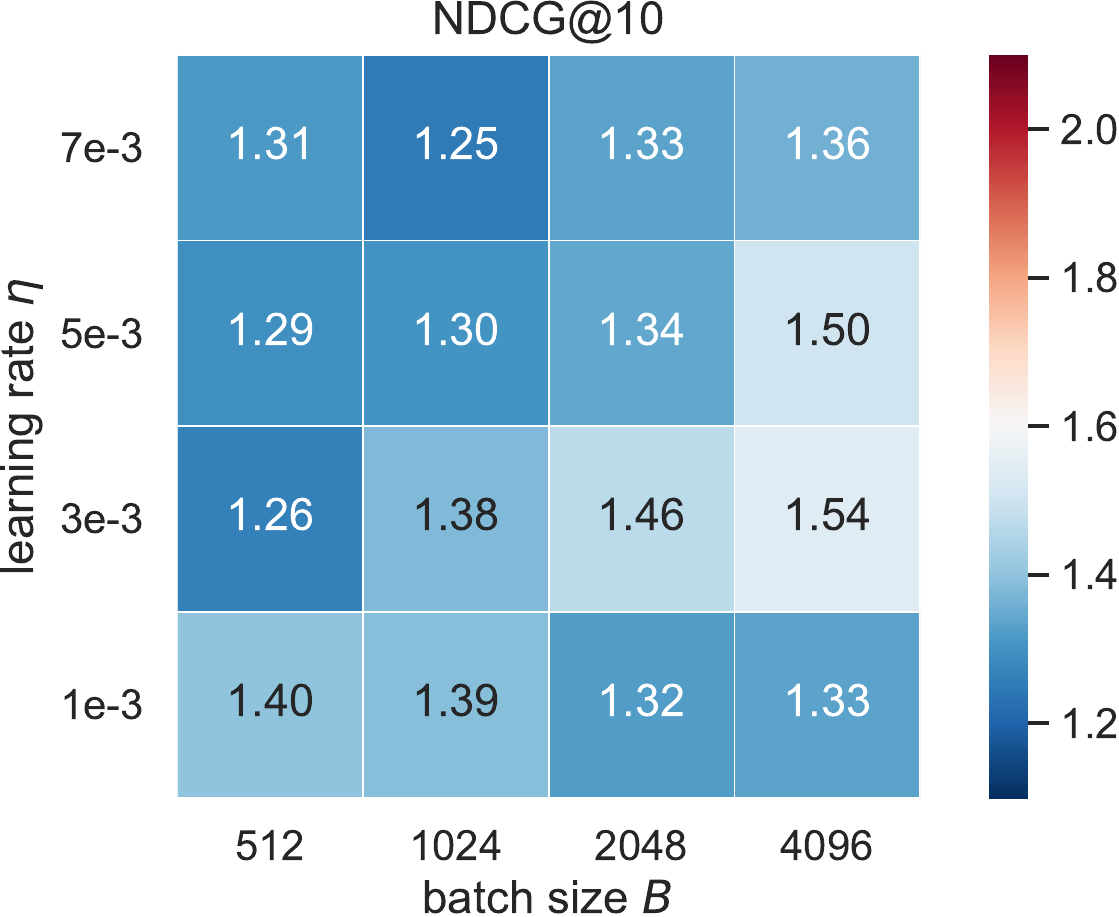}
\caption{Transformer}
\label{fig:grad_ndcg_ama_va}
\end{subfigure}
\begin{subfigure}{0.24\textwidth}
\centering
\includegraphics[height=2.7cm]{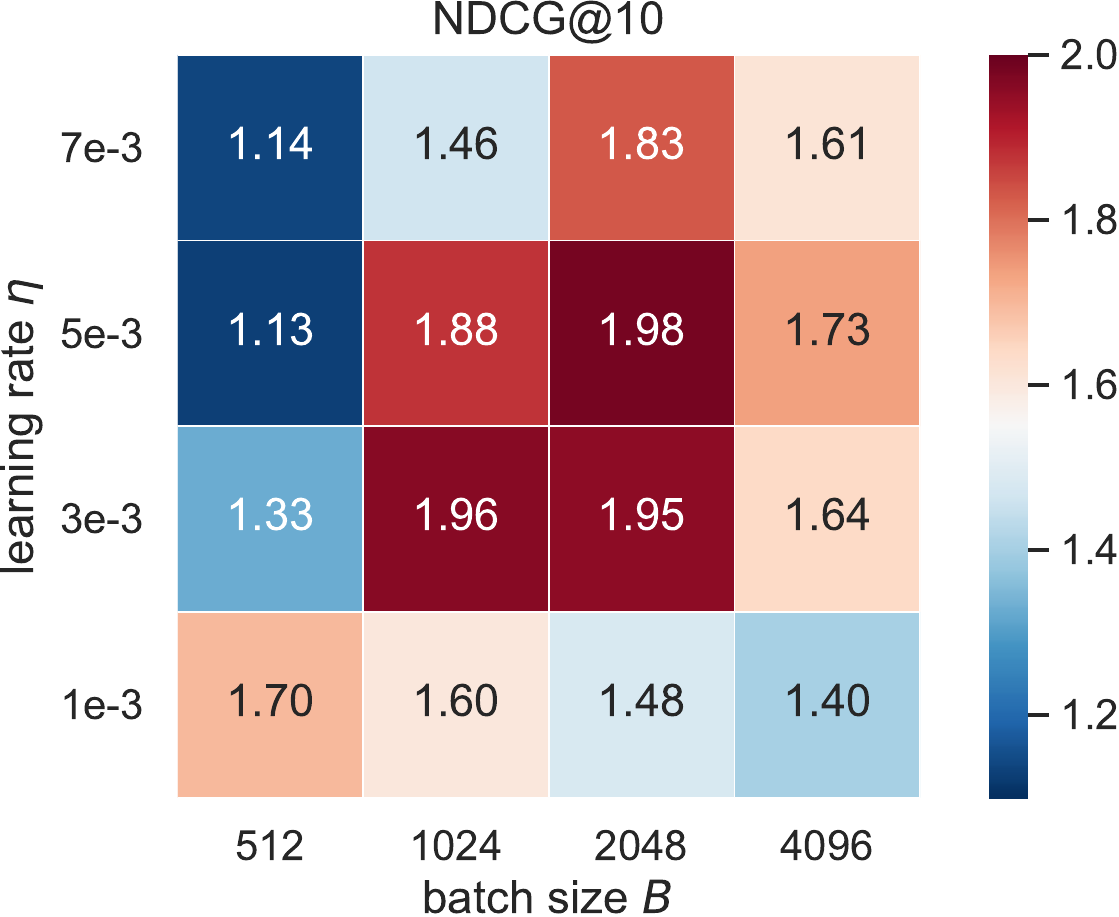}
\caption{DPFormer}
\label{fig:grad_hit_ama_dp}
\end{subfigure}
\begin{subfigure}{0.24\textwidth}
\centering
\includegraphics[height=2.7cm]{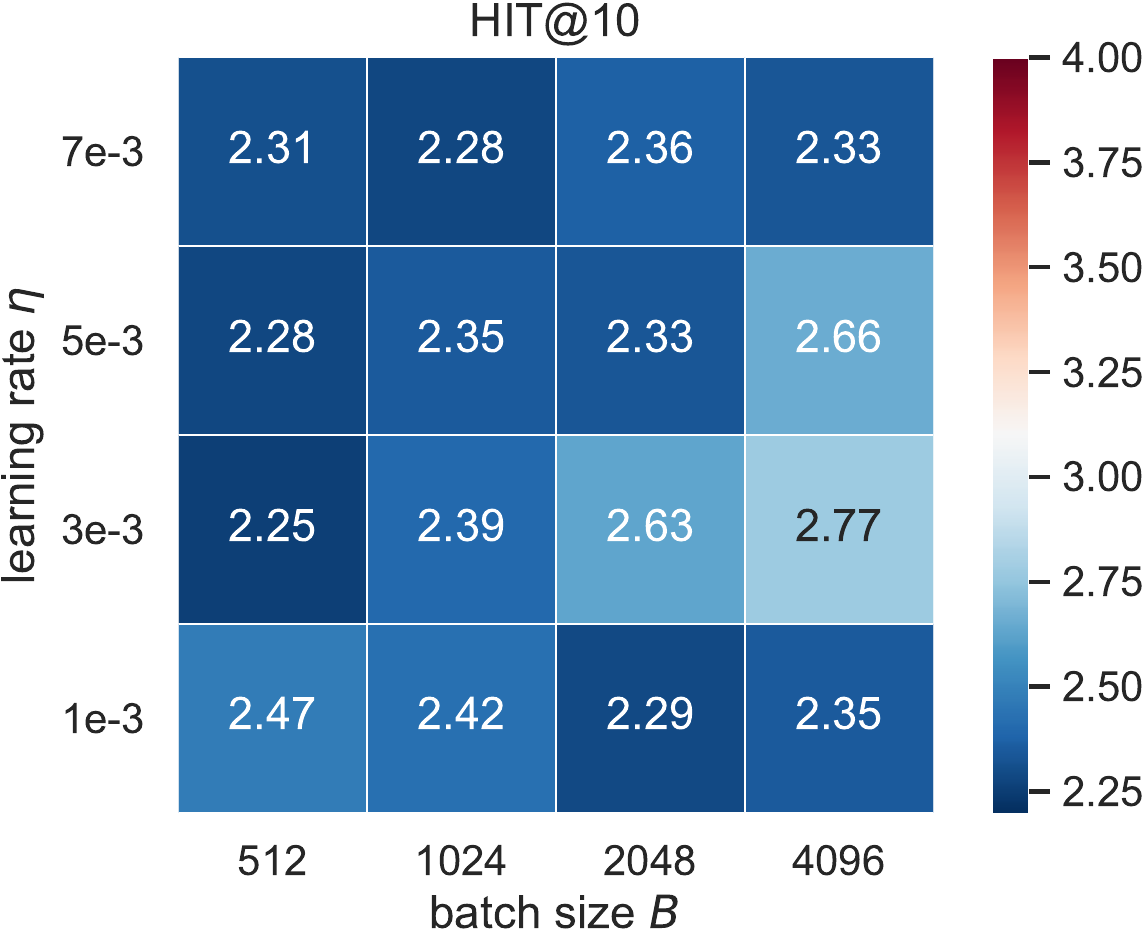}
\caption{Transformer}
\label{fig:grad_hit_ama_va}
\end{subfigure}
\caption{Results of grid search for hyperparameter tuning on Amazon with privacy budget $\varepsilon=8$.}
\label{fig:gird_exp}
\end{figure}

Figure~\ref{fig:gird_exp} shows the results of hyperparameter tuning via grid search~\footnote{Rather than run an addtional differentially private algorithm to report a noisy max (or argmax)~\cite{papernot2022hyperparameter}, we opt for this practice of directly displaying all results due to its transparency and comprehensiveness.}. For most hyperparameter configurations (especially along the main diagonal~\cite{tramerdifferentially}), our DPFormer significantly outperforms the vanilla Transformer.

\section{Conclusion}
In this paper, we identify two key challenges in learning differentially private Transformers, i.e., heavy computation overhead due to per-sample gradient clipping and attention shift due to long-tailed data distributions. We then proposed DPFormer, equipped with Phantom Clipping and Re-Attention Mechanism, to address these challenges. Our theoretical analysis shows that DPFormer can effectively correct attention shift (which leads to significant performance drops) and reduce computational cost during gradient clipping. Such analysis is further corroborated by empirical results on two real-world datasets. \textbf{Limitation.} One limitation of this work is the assumption of a trusted data curator. Another limitation is the absence of semantic interpretation for the privacy budget, given the sequential data as input. Addressing these limitations would be interesting future work.

\newpage
\appendix

\section{Related Work}
\label{apdx:related}
The most related work is \cite{li2022large}, which finetunes large language models with differential privacy. They propose Ghost Clipping, which is a technique that enables efficient per-sample gradient clipping without instantiating per-sample gradient. Our Phantom Clipping can be viewed as an extension of Ghost Clipping that additionally handles parameter sharing of the embedding layer. They also introduce the idea of \emph{effective noise multiplier} in order to explain the role of batch size in private learning. Our \emph{effective error} (Equation~\ref{eq:effectiveerror}) can be viewed as its generalization in order to account for the inherent input sparsity (i.e., only a small portion of tokens appear in one training sequence).

Another related work is \cite{anil-etal-2022-large}, which establishes a baseline for BERT-Large pretraining with DP. They introduce several strategies to help private training, such as large weight decay and increasing batch size schedule. Notably, these strategies are independent yet complementary to the methodologies utilized in this work, thereby offering potential avenues for an integrated approach.

\section{Proof of Phantom Clipping (Claim~\ref{eq:phantom})}
\label{apdx:pfphan}
\begin{proof}
For simplicity, we will omit the per-sample index $i$ throughout this proof.
From the chain rule, the per-sample gradient with respect to the embedding layer $E$ is
\begin{equation}
\begin{split}
    g_{E} &= 
    \frac{\partial \mathcal{L}}{\partial e_{s}}\cdot \frac{\partial e_{s}}{\partial E} + 
    \frac{\partial \mathcal{L}}{\partial e_{c}}\cdot \frac{\partial e_{c}}{\partial E}\\
    &=\underbrace{a_s^T \cdot \nabla e_{s}}_{g_{E}^{(1)}} + 
    \underbrace{a_c^T \cdot \nabla e_{c}}_{g_{E}^{(2)}},
\end{split}
\end{equation}
where $a_{s} \in \{0, 1\}^{L\times M}$ (or $a_{c} \in \{0, 1\}^{M\times M}$) is the one-hot encodings of the input sequence $s_i$ (or those of the candidate tokens for the output probability) in a minibatch, and $e_{s} \in \mathbb{R}^{L\times d}$ (or $e_{c} \in \mathbb{R}^{M\times d}$) be output of the (shared) embedding layer $E$ when fed into $a_{s}$ (or $a_{c}$).
Denote the first segment of the right-hand side (RHS) as $g_{E}^{(1)}$, the second segment of the RHS as $g_{E}^{(2)}$. Then we have
\begin{equation}
     \left\|g_{E}\right\|_F^2 = \left\|g_{E}^{(1)} + g_{E}^{(2)}\right\|_F^2 =  
     \left\|g_{E}^{(1)}\right\|_F^2 + \left\| g_{E}^{(2)}\right\|_F^2
     + 2\cdot \langle g_{E}^{(1)}, g_{E}^{(2)}\rangle.
     \label{eq:breakdown}
\end{equation}
Ghost Clipping~\cite{li2022large} allows us to evaluate $\left\|g_{ E}^{(1)}\right\|_F$  without instantiating $g_{E}^{(1)}$, the formula is given by
\begin{equation}
    \begin{split}
        \left\|g_{E}^{(1)}\right\|_F = \langle a_{s}a_{s}^T, \nabla e_{s} \nabla e_{s}^T \rangle.
    \end{split}
    \label{eq:phantom_p_1}
\end{equation}
Similarly, we have
\begin{equation}
    \begin{split}
        \left\|g_{E}^{(2)}\right\|_F = \langle a_{c}a_{c}^T, \nabla e_{c} \nabla e_{c}^T \rangle.
    \end{split}
    \label{eq:ghost_emb}
\end{equation}
Note that $a_{i, c}$ is the one-hot encoding of $[1, 2, 3, ..., M]$, thus $a_{i, c}$ is an Identity matrix, Equation~\ref{eq:ghost_emb} can be further simplified as
\begin{equation}
    \begin{split}
        \left\|g_{i, E}^{(2)}\right\|_F = \langle \mathbf{I}, \nabla e_{i, c} \nabla e_{i, c}^T \rangle = \sum_{i=1}^{M} \langle (\nabla e_{s})_i, (\nabla e_{s})_i \rangle =  \|\nabla e_{i, c}\|^2.
    \end{split}
    \label{eq:phantom_p_2}
\end{equation}
Note that this simplification reduces the memory footprint from $O(M^2)$ to $O(M)$, obviating the need for evaluating $e_{c} \nabla e_{c}^T$ in Equation~\ref{eq:ghost_emb}.

Therefore, computing the gradient norm of shared embedding reduces to computing $\langle g_{E}^{(1)}, g_{E}^{(2)}\rangle$ in Equation~\ref{eq:breakdown}.
\begin{equation}
\begin{split}
    \langle g_{E}^{(1)}, g_{E}^{(2)}\rangle 
    &= \langle a_{s}^T \cdot  \nabla e_{s}, a_{c}^T \cdot \nabla e_{c}\rangle \\
    &= \sum_{j=1}^{M}\sum_{k=1}^{d}\left(\sum_{i=1}^{L} (a_{s})_{ij} \cdot (\nabla e_{s})_{ik}\right)\left(\sum_{i=1}^{M} (a_{c})_{ij} \cdot (\nabla e_{c})_{ik}\right)\\
    &= \sum_{i_1=1}^{L}\sum_{i_2=1}^{M} \left(\sum_{j=1}^{M}\sum_{k=1}^{d} (a_{s})_{i_1j} \cdot (\nabla e_{s})_{i_1k} \cdot (a_{c})_{i_2j} \cdot (\nabla e_{c})_{i_2k} \right) \\
    &= \sum_{i_1=1}^{L}\sum_{i_2=1}^{M}\left(\sum_{j=1}^{M} (a_{s})_{i_1j} \cdot (a_{c})_{i_2j}\right)\left(\sum_{k=1}^{d} (\nabla e_{s})_{i_1k} \cdot (\nabla e_{c})_{i_2k}\right)\\
    &= \sum_{i_1=1}^{L}\sum_{i_2=1}^{M}\langle (a_{s})_{i_1}, (a_{c})_{i_2}\rangle \cdot \langle (\nabla e_{s})_{i_1}, (\nabla e_{c})_{i_2}\rangle\\
    &= \sum_{i_1=1}^{L}\sum_{i_2=1}^{M}[i_2 == \operatorname{onehot}^{-1}((a_{s})_{i_1})] \cdot \langle (\nabla e_{s})_{i_1}, (\nabla e_{c})_{i_2}\rangle\\
    &= \sum_{i_1=1}^{L}\langle (\nabla e_{s})_{i_1}, (a_s^T \cdot \nabla  e_{c})_{i_2}\rangle\\
    &= \langle (\nabla e_{s}), a_s^T\cdot (\nabla e_{c})\rangle.\\
\end{split}
\label{eq:phantom_p_3}
\end{equation}
Combining Equation~\ref{eq:phantom_p_1},~\ref{eq:phantom_p_2} and ~\ref{eq:phantom_p_3} yields the conclusion.
\end{proof}

\section{Proof of Error Instantiation (Claim~\ref{cl:errorinst1} and \ref{cl:errorinst2})}
\label{apdx:errorinst}
\textcolor{red}{Correction of typo: The definition of effective batch size $B_{\eff}$ in Equation~\ref{eq:effectiveerror} should be
\begin{equation}
    B_{\eff}^{\theta} = \E_{\mathcal{B}\stackrel{\tiny\text{i.i.d}}{~\sim~}\mathcal{D}^B} \left[ \sum_{i=1}^B \mathbb{I}\left[R_{\theta}(\mathcal{B}_i)\right]\right].
\end{equation}}

Taking the average over the minibatch in Equation~\ref{eq:DPSGD} can be considered noise reduction of $O(1/B)$. Fix the noise multiplier $\sigma_{dp}$. As the size of the batch increases, the amount of DP noise incorporated into the parameter decreases correspondingly. Suppose that token $i$ is absent from the current training sequence. Its input embedding will not be activated and thus will not be properly trained in this iteration, but the DP noise will be nevertheless injected into its embedding. The concept of effective error is introduced to account for this phenomenon.

\begin{proof}
    It reduces the derive the formula for effective batch size $B_{\eff}^{\theta}$ in Equation~\ref{eq:effectiveerror}.
    Recall that its definition is given by
    \begin{equation}
        B_{\eff}^{\theta} = \E_{\mathcal{B}\stackrel{\tiny\text{i.i.d}}{~\sim~}\mathcal{D}^B} \left[\sum_{i=1}^B \mathbb{I}\left[ R_{\theta}(\mathcal{B}_i)\right]\right].
    \end{equation}
    For each layer parameterized by $W$ within the Transformer block, its effective batch size is $B_{\eff}^{W} = B$, since $R_{W}(\mathcal{B}_i) = 1$.

    For the embedding layer $E$, its effective batch size is
    \begin{equation}
    \begin{split}
         B_{\eff}^{E_i} &=  \E_{\mathcal{B}\stackrel{\tiny\text{i.i.d}}{~\sim~}\mathcal{D}^B} \left[\sum_{j=1}^B \mathbb{I}\left[ R_{E_i}(\mathcal{B}_j)\right]\right]\\
         &= \sum_{j=1}^B  \E_{\mathcal{B}_j\stackrel{\tiny\text{i.i.d}}{~\sim~}\mathcal{D}} \left[ \mathbb{I}\left[ R_{E_i}(\mathcal{B}_j)\right]\right] ~~~~\text{(Linearity of Expectation)}\\
         &= \sum_{j=1}^B  \E_{\mathcal{B}_j\stackrel{\tiny\text{i.i.d}}{~\sim~}\mathcal{D}} \left[ \mathbb{I}\left[ {\rm token}~i\in \mathcal{B}_j \right]\right]\\
         &= \sum_{j=1}^B  p_i\\
         &= B\cdot p_i,\\
    \end{split}
    \end{equation}
    where $p_i$ is the frequency of token $i$ (i.e., the probability of token $i$'s occurrence in data). 
\end{proof}
\begin{remark}
    Note that $p_i$, as the frequency statistics, can be obtained with high accuracy with tiny privacy budget. For simplicity,  we will assume $p_i$ is publicly known in this work.
\end{remark}

\section{Proof of Error Propagation}
\subsection{Proof of Linear Transformation (Equation~\ref{eq:linearnpn})}
\label{apdx:npnproof}
\begin{lemma}
    Let $X$, $Y$ be two independent random variables. Let $Z=XY$, then the variance of $Z$ can be expressed as
    \begin{equation}
        \begin{split}
            \var[Z] = \var[XY] =  \E[X^2]\E[Y^2] - \E[XY]^2.
        \end{split}
    \end{equation}
    \label{le:prod_var}
\end{lemma}
Lemma\ref{le:prod_var} directly implies Equation~\ref{eq:linearnpn} as follows.
\begin{proof}
    Suppose the linear transformation is given by $X^{(l)}=X^{(l-1)}W$, then we have
    \begin{equation}
    \begin{split}
        \var[X^{(l)}] &= \E[(X^{(l-1)})^2]\E[W^2] - \E[X^{(l-1)}W]^2\\
        &= (\E[X^{(l-1)}]^2 + \var[X^{(l-1)}])(\E[W]^2 + \var[W])- \E[X^{(l-1)}]^2\E[W]^2\\
        &= \var[X^{(l-1)}]\var[W] + \E[X^{(l-1)}]^2\var[W] + \E[W]^2\var[X^{(l-1)}].\\
    \end{split}
    \end{equation}
\end{proof}
\subsection{Proof of Non-linear Transformation (Equation~\ref{eq:nonprop})}
\begin{figure}[h]
\vspace{-0.5cm}
\centering
\setlength{\abovecaptionskip}{0.cm}
\includegraphics[width=10cm]{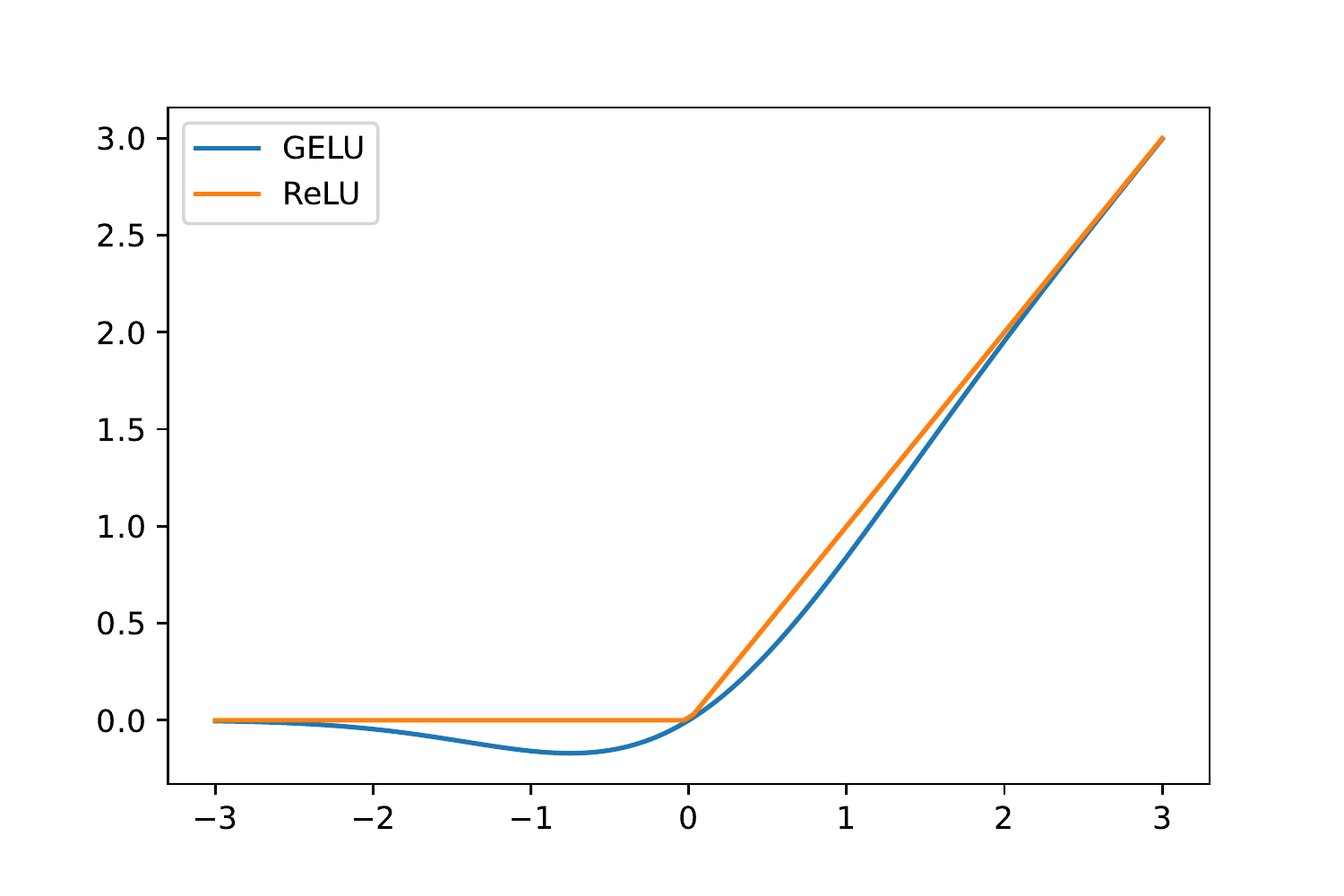}

\caption{GELU activation and ReLU activation.}
\label{apdx:fig:gelu}

\end{figure}

\begin{lemma}
\label{le:max_var}
    Let $X_1$, $X_2$ be two independent Gaussian random variables, where $X_i\sim \mathcal{N}(\mu_i, \sigma_i),i=1,2$. Let $Z=\operatorname{max}(X_1, X_2)$.
    \begin{equation}
        \begin{split}
            &\E[Z] = \mu_1 \Phi(\gamma) + \mu_2\Phi(-\gamma) + \nu\phi(\gamma)\\
            &\E[Z^2] = (\mu_1^2+\sigma_1^2) \Phi(\gamma) + (\mu_2^2 + \sigma_2^2)\Phi(-\gamma) + (\mu_1 + \mu_2)\nu\phi(\gamma),
        \end{split}
        \label{eq:max_var}
    \end{equation}
    where $\Phi(\cdot)$ is the cumulative density function (CDF) of the standard Gaussian distribution, $\phi(\cdot)$ is the probability density function (PDF) of the standard Gaussian distribution, $\nu$ = $\sqrt{\sigma_1^2 + \sigma_2^2}$, and $\gamma=(\mu_1-\mu_2)/\nu$.
\end{lemma}
Lemma\ref{le:max_var} directly implies Equation~\ref{eq:nonprop} as follows.
\begin{proof}
    Let $X^{(l)}=\operatorname{ReLU}\left(X^{(l-1)}\right)$ be the ReLU activation. Substitute $\mu_1=\E[X^{(l-1)}], \sigma_2 =\var[X^{(l-1)}], \mu_2=\sigma_2 = 0$ into Equation~\ref{eq:max_var}. Leveraging $\var[X^{(l)}]=\E[(X^{(l)})^2] -\E[X^{(l)}]^2$  yields the conclusion.
\end{proof}
\begin{remark}
    \textbf{GELU activation.} GELU function is another widely used activation function within Transformer models. GELU can be viewed as a smooth version of ReLU (see Figure~\ref{apdx:fig:gelu}), where their forward propagation is similar, and the major distinction lies in the numerical behavior of backpropagation. Since error propagation is only concerned with forward propagation behavior, we can also use Equation~\ref{eq:nonprop} to approximate the variance of GELU output. Table~\ref{apdx:tb:analytic} shows the analytic error propagation for ReLU and GELU activation, compared with the sampling-based results.
\end{remark}

\begin{table}[h]
\scriptsize
\centering
\caption{Analytic error propagation for ReLU and GELU activation.}
\begin{tabular}{@{}c|c|cccccc|c@{}}
\toprule
\multirow{2}{*}{Input}      & \multirow{2}{*}{Activation} & \multicolumn{6}{c|}{Sampling-based}                       & \multirow{2}{*}{Analytic} \\
                            &                             & 10      & 100     & 1000    & 10000   & 100000  & 1000000 &                           \\ \midrule\midrule
\multirow{2}{*}{$\mathcal{N}(0, 0.01)$} & \textsc{ReLU}                        & 4.08e-6 & 3.60e-5 & 3.93e-5 & 3.45e-5 & 3.40e-5 & 3.40e-5 & \multirow{2}{*}{\textbf{3.40e-5}}  \\
                            & \textsc{GELU}                         & 2.48e-5 & 2.69e-5 & 2.72e-5 & 2.57e-5 & 2.50e-0 & 2.49e-5 &                           \\ \midrule
\multirow{2}{*}{$\mathcal{N}(0, 0.1)$}  & \textsc{ReLU}                        & 0.0030  & 0.0031  & 0.0037  & 0.0034  & 0.0035  & 0.0034  & \multirow{2}{*}{\textbf{0.0034}}   \\
                            & \textsc{GELU}                        & 0.0030  & 0.0025  & 0.0027  & 0.0025  & 0.0026  & 0.0025  &                           \\ \midrule
\multirow{2}{*}{$\mathcal{N}(0, 1)$}    & \textsc{ReLU}                        & 0.5299  & 0.2361  & 0.3649  & 0.3451  & 0.3387  & 0.3418  & \multirow{2}{*}{\textbf{0.3408}}   \\
 & \textsc{GELU}  & 0.5525  & 0.2306  & 0.3719  & 0.3506  & 0.3433  & 0.3467  &  \\ \bottomrule
\end{tabular}
\label{apdx:tb:analytic}
\end{table}

\section{Experimental Details}
\label{apdx:exp}
\subsection{Datasets}
\label{apdx:exp:dataset}
\textbf{MovieLens.} The MovieLens dataset~\cite{harper2015movielens} is often used in the development and evaluation of collaborative filtering algorithms, which are used to make personalized recommendations based on user behavior. It is a benchmark dataset in the field of recommender systems due to its size, longevity, and richness of user-item interactions. We use the version (MovieLens-1M) that includes 1 million user behaviors.

\textbf{Amazon.} A series of datasets introduced in~\cite{mcauley2015image}, comprising large corpora of product reviews crawled from
Amazon.com. Top-level product categories on Amazon are
treated as separate datasets. We consider the ‘Games.’ category. This dataset is notable for its high sparsity and variability.

We follow \cite{kang2018self} for the data preprocessing. We use timestamps to determine the sequence order of
actions. Each user is associated with a training sequence (i.e., his chronological behavior). We discard users and items with fewer than five related actions. For data partitioning, the last token of each sequence is left for testing.

It is worth noting that since each user is exclusively associated with exactly one training sample (sequence) in the training data, the DP guarantee we provide is user-level. That is, removing all information pertaining to a specific user  yields an indistinguishable model.

\subsection{Model Architecture}
\label{apdx:exp:preprocessing}
We use the standard Transformer encoder described in~\cite{vaswani2017attention}. The model dimension is set to 64. The number of heads in the Attention Mechanism is set to 1. The number of transformer blocks is set to 2. Our model adopts a learned (instead of fixed) positional embedding.

\subsection{Hyperparameters}
The number of epochs is set to 100. The batch size is chosen from $\{256, 512, 1024, 2048, 4096\}$. The learning rate is chosen from $\{10^{-3}, 3\times10^{-3}, 5\times10^{-3},  7\times10^{-3},  9\times10^{-3}\}$. The dropout rate is 0.2 for MovieLens and 0.5 for Amazon (due to its high sparsity). We use the Adam optimizer with a weight decay of $10^{-5}$.

\end{document}